\documentclass[10pt,twocolumn,letterpaper]{article}

\usepackage{iccv}
\usepackage{times}
\usepackage{epsfig}
\usepackage{graphicx}
\usepackage{amsmath}
\usepackage{amssymb}
\usepackage[accsupp]{axessibility}  % Improves PDF readability for those with disabilities.

\usepackage{bm}
\usepackage{multirow}

\usepackage[english]{babel}

\usepackage{algorithm} %format of the algorithm
\usepackage{algpseudocode} %format of the algorithm
\usepackage{booktabs}
\usepackage{caption}
\usepackage{subcaption}
\usepackage{verbatim}
\usepackage{gensymb}
\usepackage{rotating}
\usepackage{marvosym}

\makeatletter
\@namedef{ver@everyshi.sty}{}
\makeatother
\usepackage{tikz}

\makeatletter
\@namedef{ver@everyshi.sty}{}
\makeatother
\usepackage{pgffor}

\usepackage{xspace}
\usepackage{enumitem}

\usepackage[labelformat=simple]{subcaption}

\newcommand\blfootnote[1]{%	
  \begingroup
  \renewcommand\thefootnote{}\footnote{#1}%
  \addtocounter{footnote}{-1}%
  \endgroup
}

\usepackage[pagebackref=true,breaklinks=true,letterpaper=true,colorlinks=true,bookmarks=false]{hyperref}

\iccvfinalcopy % *** Uncomment this line for the final submission

\def\iccvPaperID{3358} % *** Enter the ICCV Paper ID here

\begin{document}

%%%%%%%%% TITLE
\title{PyMAF: 3D Human Pose and Shape Regression with Pyramidal Mesh Alignment Feedback Loop}

\newcommand{\aufont}{\fontsize{11}{12}\selectfont}
\newcommand{\infont}{\fontsize{10}{11}\selectfont}

\author{
\aufont Hongwen Zhang\textsuperscript{$\mathsection\ddagger$*}, Yating Tian\textsuperscript{$\dagger$*}, Xinchi Zhou\textsuperscript{$\natural$}, Wanli Ouyang\textsuperscript{$\natural$}, Yebin Liu\textsuperscript{$\ddagger$}, Limin Wang\textsuperscript{$\dagger$ \Letter}, Zhenan Sun\textsuperscript{$\mathsection$ \Letter}\\
\infont\textsuperscript{$\mathsection$}CRIPAC, NLPR, Institute of Automation, Chinese Academy of Sciences, China\\
\infont\textsuperscript{$\dagger$}State Key Laboratory for Novel Software Technology, Nanjing University, China\\
% \infont\textsuperscript{$\natural$}The University of Sydney, SenseTime Computer Vision Research Group, Australia\\
\infont\textsuperscript{$\natural$}The University of Sydney, Australia~~\textsuperscript{$\ddagger$}Department of Automation, Tsinghua University, China\\
{\tt\small
\{hongwen.zhang@cripac,znsun@nlpr\}.ia.ac.cn~~\{yatingtian@smail.,lmwang@\}nju.edu.cn}\\
{\tt\small
\{xinchi.zhou1,wanli.ouyang\}@sydney.edu.au~~liuyebin@mail.tsinghua.edu.cn}
}

\maketitle

%%%%%%%%% ABSTRACT
\begin{abstract}
Regression-based methods have recently shown promising results in reconstructing human meshes from monocular images. By directly mapping raw pixels to model parameters, these methods can produce parametric models in a feed-forward manner via neural networks. However, minor deviation in parameters may lead to noticeable misalignment between the estimated meshes and image evidences. To address this issue, we propose a Pyramidal Mesh Alignment Feedback (PyMAF) loop to leverage a feature pyramid and rectify the predicted parameters explicitly based on the mesh-image alignment status in our deep regressor. In PyMAF, given the currently predicted parameters, mesh-aligned evidences will be extracted from finer-resolution features accordingly and fed back for parameter rectification. To reduce noise and enhance the reliability of these evidences, an auxiliary pixel-wise supervision is imposed on the feature encoder, which provides mesh-image correspondence guidance for our network to preserve the most related information in spatial features. The efficacy of our approach is validated on several benchmarks, including Human3.6M, 3DPW, LSP, and COCO, where experimental results show that our approach consistently improves the mesh-image alignment of the reconstruction. The project page with code and video results can be found at \href{https://hongwenzhang.github.io/pymaf}{https://hongwenzhang.github.io/pymaf}.
\end{abstract}

\blfootnote{*: Equal contribution. \Letter: Corresponding authors.}

%%%%%%%%% BODY TEXT
\vspace{-1mm}

\section{Introduction}\label{sec:introduction}

\begin{figure}[t]
	\raggedright
    \begin{subfigure}[b]{0.14\textwidth}
		\includegraphics[height=32mm]{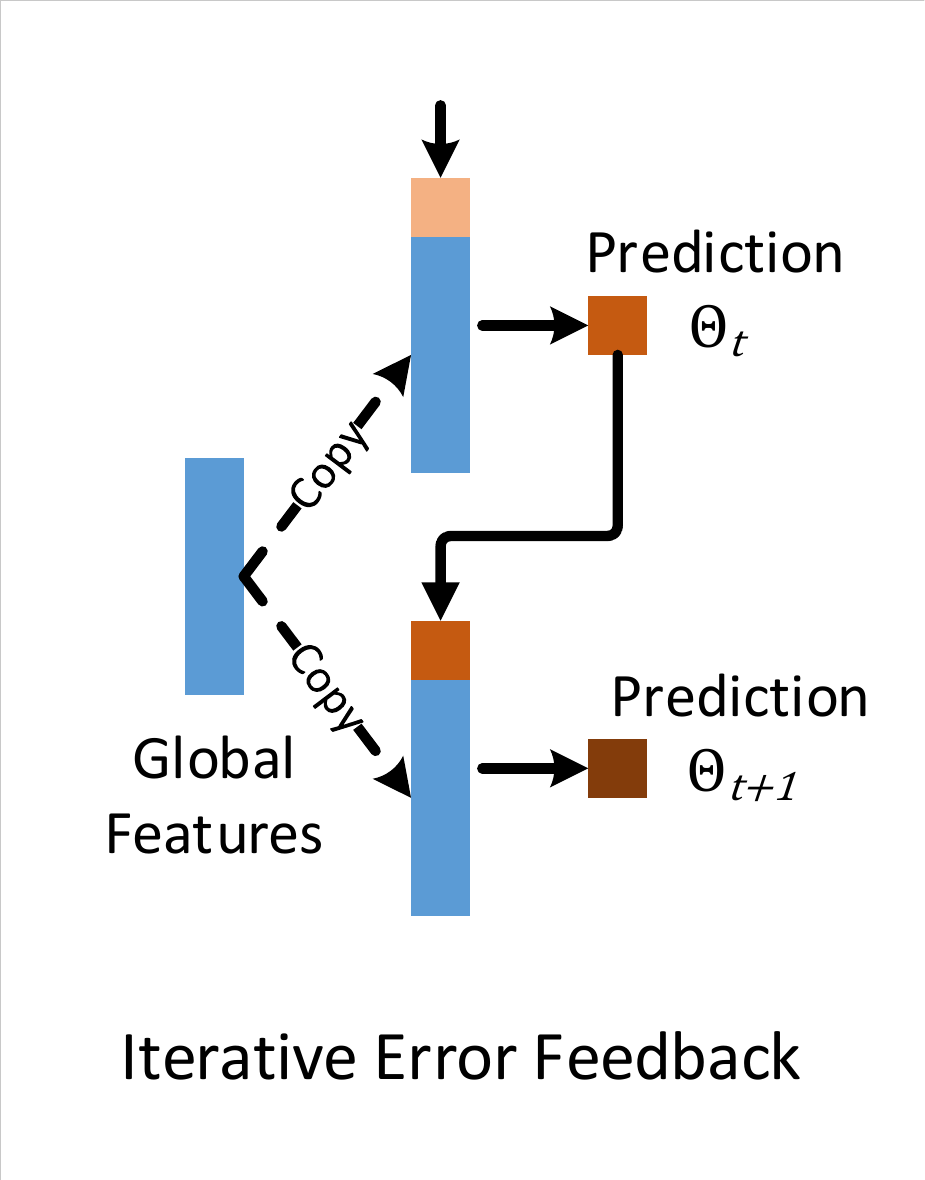}
		\caption{}
		\label{fig:ief}
    \end{subfigure}
    \begin{subfigure}[b]{0.19\textwidth}
		\includegraphics[height=32mm]{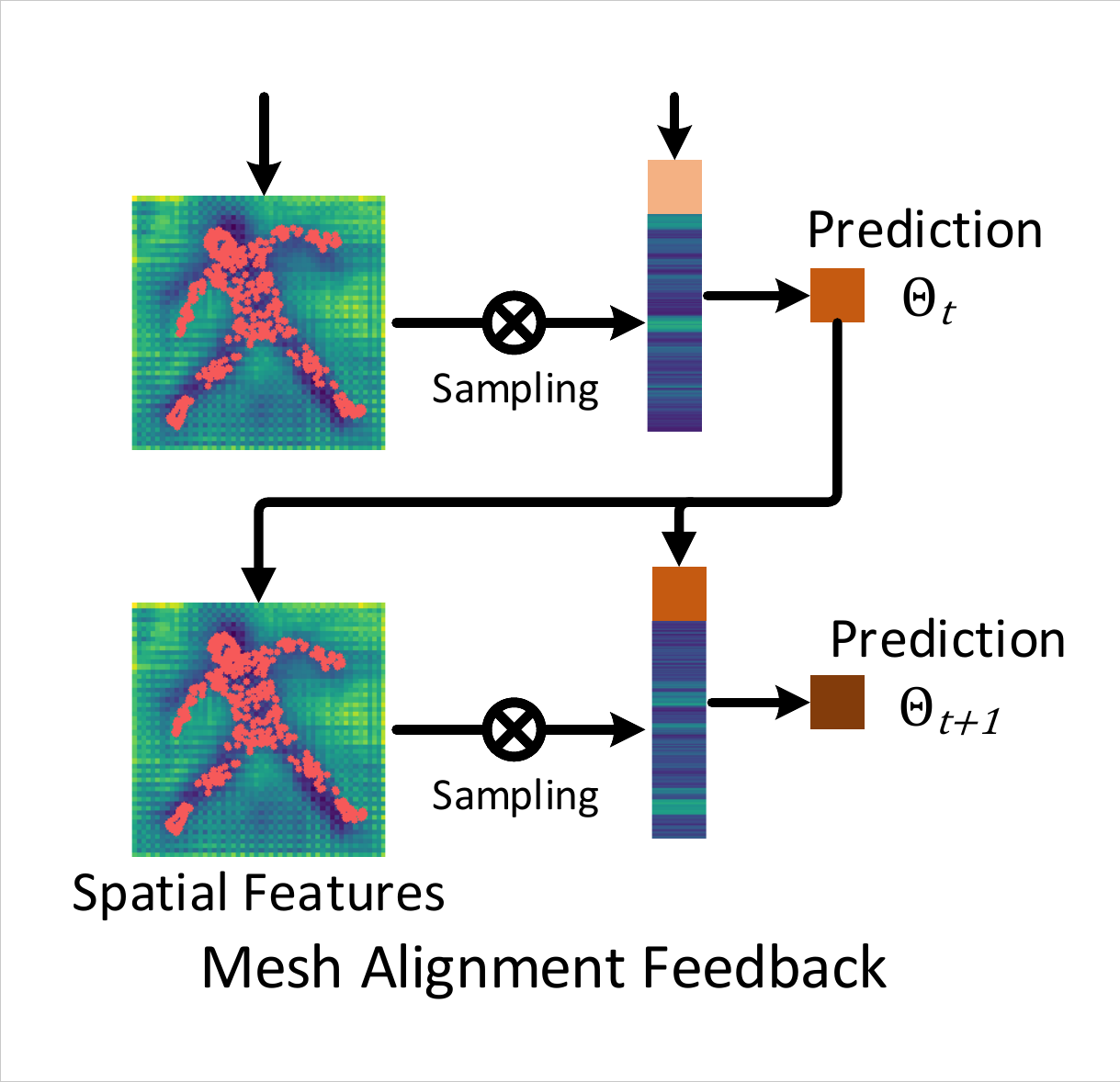}
		\caption{}
		\label{fig:maf}
    \end{subfigure}
    \begin{subfigure}[b]{0.13\textwidth}
		\includegraphics[height=32mm]{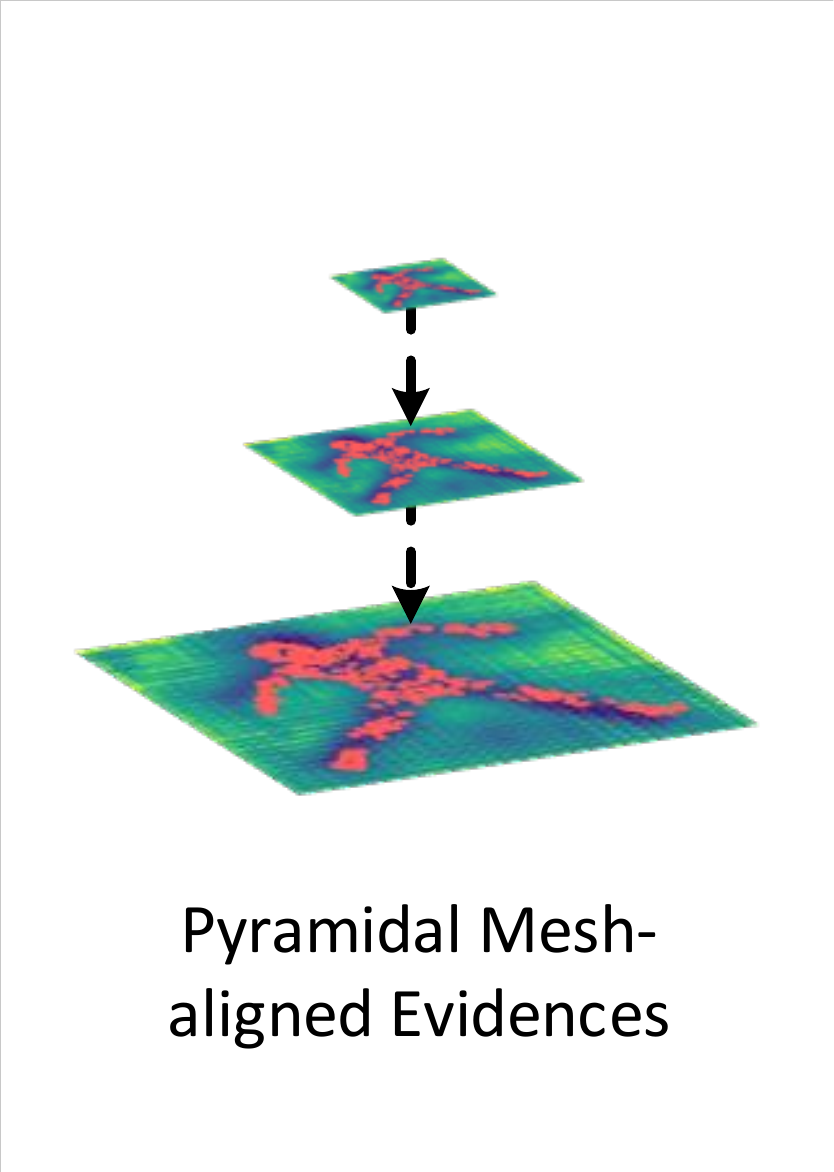}
		\caption{}
		\label{fig:pyramid}
    \end{subfigure}
    \begin{subfigure}[b]{0.40\textwidth}
		\includegraphics[height=40mm]{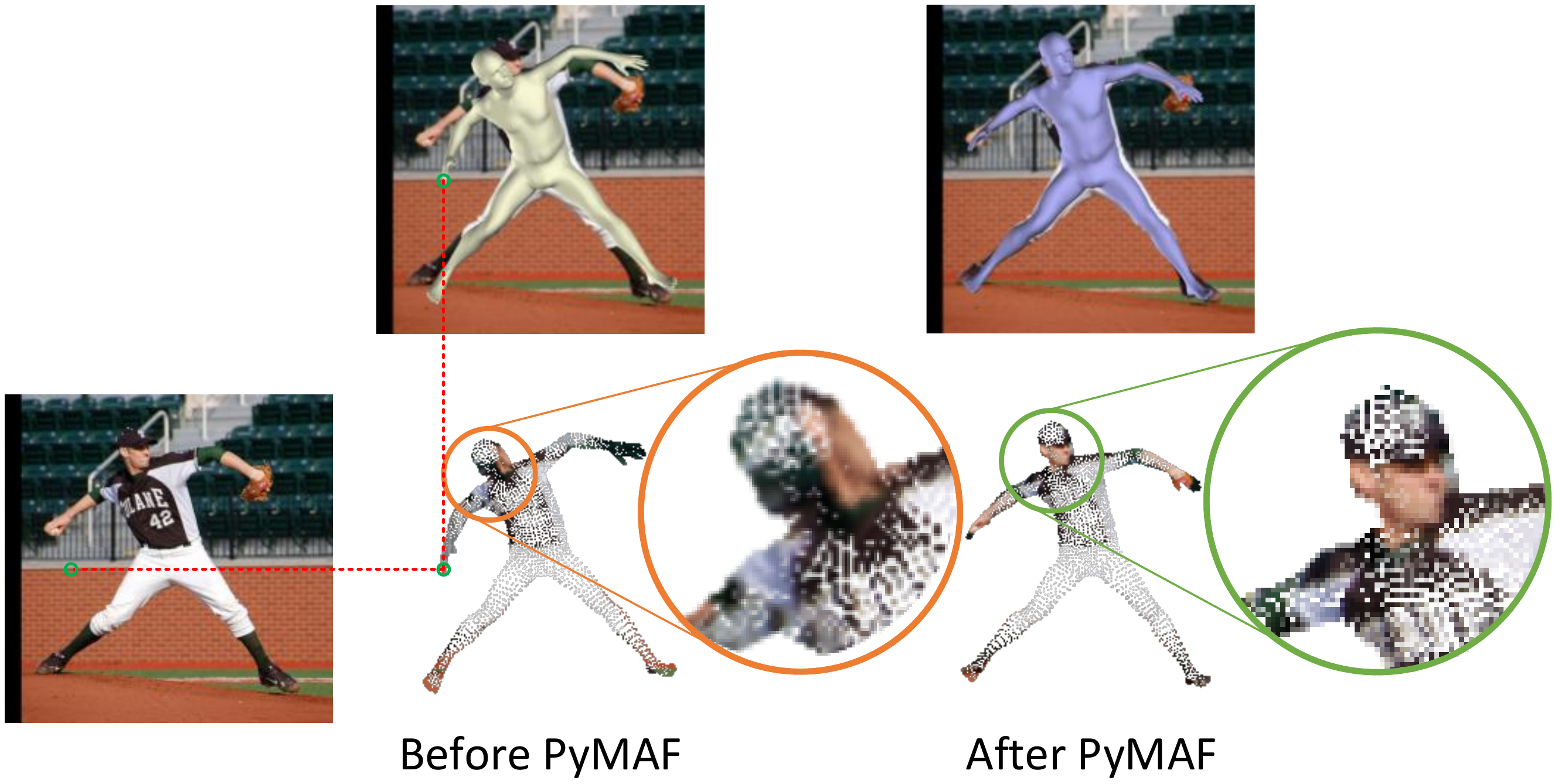}
		\caption{}
		\label{fig:alignsmpl}
    \end{subfigure}
    \vspace{-3mm}
	\caption{Illustration of our main idea. (a) The commonly-used iterative error feedback. (b) Our mesh alignment feedback. (c) Mesh-aligned evidences extracted from a feature pyramid. (d) Our approach PyMAF improves the mesh-image alignment of the estimated mesh.}
	\vspace{-5mm}
	\label{fig:teaser}
\end{figure}

Aiming at the same goal of producing natural and well-aligned results, two different paradigms for human mesh recovery have been investigated in the research community.
Optimization-based methods~\cite{bogo2016keep,lassner2017unite,zanfir2018monocular} explicitly fit the models to 2D evidences, which can typically produce results with accurate mesh-image alignments but tend to be slow and sensitive to the initialization.
Alternatively, regression-based ones~\cite{kanazawa2018end,pavlakos2018learning,kolotouros2019convolutional,kolotouros2019learning} suggest to directly predict model parameters from images, which have shown very promising results, and yet still suffer from the coarse alignment between predicted meshes and image evidences.

For parametric models like SMPL~\cite{loper2015smpl}, the joint poses are represented as the relative rotations with respect to their parent joints, which means that minor rotation errors accumulated along the kinematic chain can result in noticeable drifts of joint positions.
In order to generate well-aligned results, optimization-based methods~\cite{bogo2016keep,lassner2017unite} design data terms in the objective function so that the alignment between the projection of meshes and 2D evidences can be explicitly optimized.
Similar strategies are also exploited in regression-based methods~\cite{kanazawa2018end,pavlakos2018learning,kolotouros2019convolutional,kolotouros2019learning} to impose 2D supervisions upon the projection of estimated meshes in the training procedure.
However, during testing, these deep regressors either are open-loop or simply include an Iterative Error Feedback (IEF) loop~\cite{kanazawa2018end} in their architectures.
As shown in Fig.~\ref{fig:ief}, IEF reuses the same global features in its feedback loop, making its regressors hardly perceive the mesh-image misalignment in the inference phase.

As suggested in previous works~\cite{ronneberger2015u,newell2016stacked,lin2017feature,sun2019deep}, neural networks tend to retain high-level information and discard detailed local features when reducing the spatial size of feature maps.
To leverage spatial information in the regression networks, it is essential to extract pixel-wise contexts for fine-grained perception.
Several attempts have been made to leverage pixel-wise representations such as part segmentation~\cite{omran2018neural} or dense correspondences~\cite{xu2019denserac,zhang2020learning} in their regression networks.
Though these regressors take pixel-level evidences into consideration, it is still challenging for them to learn structural priors and get hold of spatial details at the same time based merely on high-resolution contexts.

Motivated by above observations, we design a Pyramidal Mesh Alignment Feedback (PyMAF) loop in our regression network to exploit multi-scale contexts for better mesh-image alignment of the reconstruction.
The central idea of our approach is to correct parametric deviations explicitly and progressively based on the alignment status.
In PyMAF, mesh-aligned evidences will be extracted from spatial features according to the 2D projection of the estimated mesh and then fed back to regressors for parameter updates.
% These mesh-aligned evidences are position-sensitive and rich of spatial cues, which greatly help to refine the mesh so that it can be better aligned with the observations.
As illustrated in Fig.~\ref{fig:teaser}, the mesh alignment feedback loop can take advantage of more informative features for parameter correction in comparison with the commonly used iterative error feedback loop~\cite{kanazawa2018end,carreira2016human}.
To leverage multi-scale contexts, mesh-aligned evidences are extracted from a feature pyramid so that coarse-aligned meshes can be corrected with large step sizes based on the lower-resolution features.
Moreover, to enhance the reliability of the spatial cues, an auxiliary task is imposed on the highest-resolution feature to infer pixel-wise dense correspondences, which provides guidance for the image encoder to preserve the mesh-image alignment information.
% between the foreground pixels on images and the 3D vertices on meshes.
% Such dense correspondences are pixel-aligned with input images, providing guidance for the image encoder to preserve the mesh-image alignment information in the high-resolution features.
The contributions of this work can be summarized as follows:
\begin{itemize}[leftmargin=*]
\itemsep0em 
    \item A mesh alignment feedback loop is introduced for regression-based human mesh recovery, where mesh-aligned evidences are exploited to correct parametric errors explicitly so that the estimated meshes can be better-aligned with input images.
    \item A feature pyramid is further incorporated with the mesh alignment feedback loop so that the regression network can leverage multi-scale alignment contexts.
    \item An auxiliary pixel-wise supervision is imposed on the image encoder so that its spatial features can be more informative and the mesh-aligned evidences can be more relevant and reliable.
    % \item Our approach achieves competitive results on 3D pose benchmarks and improves the mesh-image alignment of the reconstructed models over previous optimization- and regression-based methods.
\end{itemize}

\section{Related Work}\label{sec:related_work}

\subsection{Human Pose and Shape Recovery}

\textbf{Optimization-based Approaches.}
Pioneering work in this field mainly focus on the optimization process of fitting parametric models (\eg, SCAPE~\cite{anguelov2005scape} and SMPL~\cite{loper2015smpl}) to 2D observations such as keypoints and silhouettes~\cite{sigal2008combined,guan2009estimating,bogo2016keep}.
In their objective functions, prior terms are designed to penalize the unnatural shape and pose, while data terms measure the fitting errors between the re-projection of meshes and 2D evidences.
Based on this paradigm, different updates have been investigated to incorporate information such as 2D/3D body joints~\cite{bogo2016keep,lightcap2021}, silhouettes~\cite{lassner2017unite,huang2017towards}, part segmentation~\cite{zanfir2018monocular} in the fitting procedure.
Despite the well-aligned results obtained by these optimization-based methods, their fitting process tends to be slow and sensitive to initialization.
Recently, Song~\etal~\cite{song2020human} exploit the learned gradient descent in the fitting process.
Though this solution leverages rich 2D pose datasets and alleviates many issues in traditional optimization-based methods, it still relies on the accuracy of 2D poses and breaks the end-to-end learning.
Alternatively, our solution supports the end-to-end learning but can also leverage rich 2D datasets thanks to the progress (\eg, SPIN~\cite{kolotouros2019learning} and EFT~\cite{joo2020exemplar}) in the generation of more precise pseudo 3D ground-truth for 2D datasets.

\textbf{Regression-based Approaches.}
Alternatively, taking advantage of the powerful nonlinear mapping capability of neural networks, recent regression-based approaches~\cite{kanazawa2018end,pavlakos2018learning,omran2018neural,kolotouros2019learning,choutas2020monocular,jiang2020coherent,choi2020pose2mesh} have made significant advances in predicting human models directly from monocular images.
These deep regressors take 2D evidences as input and learn model priors implicitly in a data-driven manner under different types of supervision signals~\cite{tung2017self,kanazawa2018end,pavlakos2019texturepose,rong2019delving,doersch2019sim2real,zanfir2020weakly,kundu2020appearance} during the learning procedure.
To mitigate the learning difficulty of the regressor, different network architectures have also been designed to leverage proxy representations such as silhouette~\cite{pavlakos2018learning,varol2018bodynet}, 2D/3D joints~\cite{tung2017self,pavlakos2018learning,moon2020i2l}, segmentation~\cite{omran2018neural,rueegg2020chained} and dense correspondences~\cite{xu2019denserac,zhang2020learning}.
Such strategies can benefit from synthetic data~\cite{xu2019denserac,sengupta2020synthetic} and the progress in the estimation of proxy representations~\cite{cao2019openpose,alp2018densepose,sun2019deep,wu2021graph}.
Despite the effectiveness of these modules, the quality of proxy representations becomes the bottleneck for the reconstruction task, which may also block the end-to-end learning of the deep regressor.
Moreover, though supervision signals are imposed on the projection of the estimated models to penalize the fitting misalignment during the training of deep regressors, their architectures can hardly perceive the misalignment during the inference phase.
In comparison, the proposed PyMAF is close-loop for both training and inference, which enables a feedback loop in our deep regressor to leverage spatial evidences for better mesh-image alignment of the estimated human models.
Our work focuses on the design of regressor architectures and can also provides a better regressor for those approaches using post-processing~\cite{guler2019holopose} or the work on pseudo ground-truth generation~\cite{kolotouros2019learning,joo2020exemplar}.

Directly regressing model parameters from images is very challenging even for neural networks.
Existing methods have also offered non-parametric solutions to reconstruct human body models in non-parametric representations.
Among them, volumetric representation~\cite{varol2018bodynet,zheng2019deephuman}, implicit function~\cite{saito2019pifu,zheng2021deepmulticap}, mesh vertices~\cite{kolotouros2019convolutional,lin2020end}, and position maps~\cite{yao2019densebody,zhang2020object,zeng20203d} have been adopted as regression targets.
Using non-parametric representations as the regression targets is more readily to leverage high-resolution features but needs further processing to retrieve parametric models from the outputs.
Besides, solely using high-resolution features makes the algorithms more sensitive to occlusions without additional structure priors.
In our solution, the deep regressor makes use of spatial features at multiple scales for high-level and fine-grained perception and produces parametric models directly with no further processing required.

\subsection{Iterative Fitting in Regression Tasks}
Strategies of incorporating fitting processes along with the regression have also been investigated in the literature.
For human model reconstruction, Kolotouros~\etal~\cite{kolotouros2019learning} combine an iterative fitting procedure to the training procedure in order to generate preciser ground truth for better supervision.
To refine the estimated meshes during both training and inference phases, several attempts have been made to deform human meshes so that they can be aligned with the intermediate estimations such as depth maps~\cite{zhu2019detailed}, part segmentation~\cite{zanfir2020neural}, and dense correspondences~\cite{guler2019holopose}.
These approaches adopt intermediate estimations as the fitting goals and hence rely on their quality.
In contrast, our approach uses the currently estimated meshes to extract deep features for refinement, which not only behaves symmetrically for both training and inference but also enables fully end-to-end learning of the deep regressor.

To put our approach in a broader view, there have been remarkable efforts made to involve iterative fitting strategies in other computer vision tasks, including facial landmark localization~\cite{xiong2013supervised,trigeorgis2016mnemonic}, human/hand pose estimation~\cite{oberweger2019generalized,carreira2016human}, \etc.
For generic objects, Pixel2Mesh~\cite{wang2018pixel2mesh} progressively deforms an initial ellipsoid by leveraging perceptual features.
Following the spirit of these works, we exploit new strategies to extract fine-grained evidences and contribute novel solutions in the context of human mesh recovery.

\begin{figure*}[t]
	\begin{center}
		\includegraphics[height=55mm]{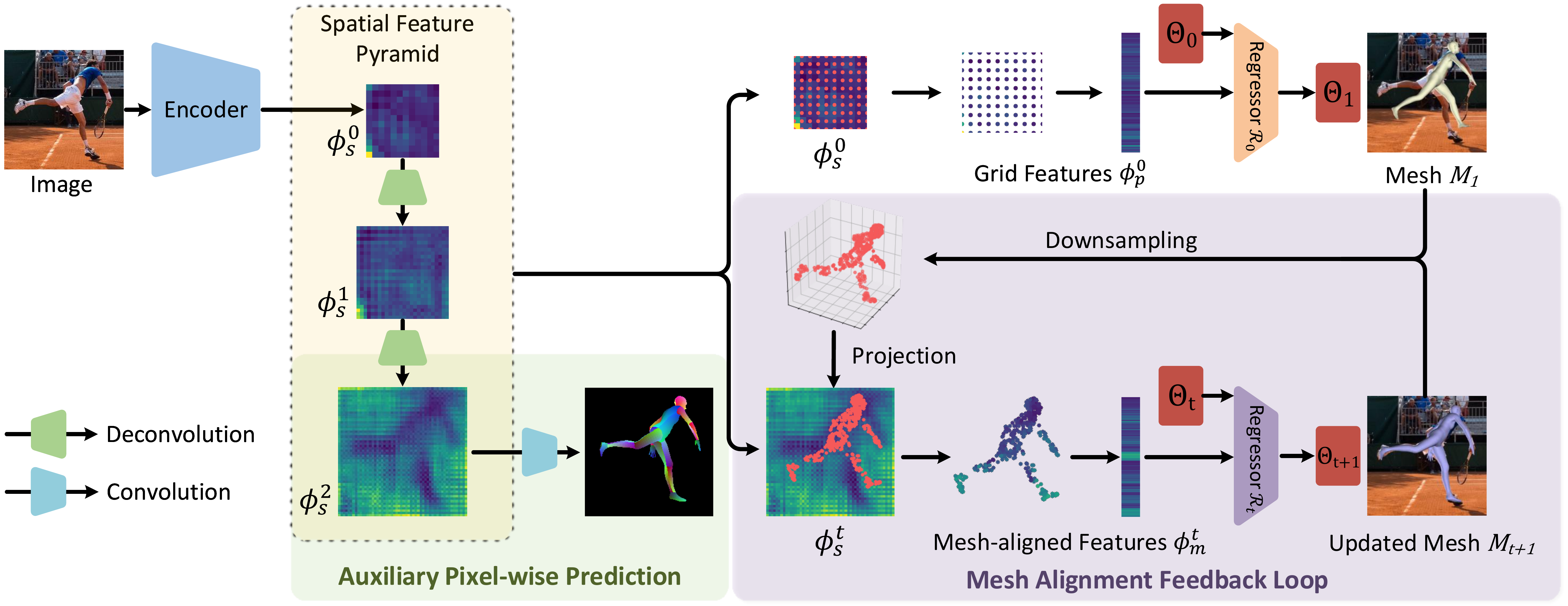}
		\caption{Overview of the proposed Pyramidal Mesh Alignment Feedback (PyMAF). PyMAF leverages a feature pyramid and enables an alignment feedback loop in our network. Given a coarse-aligned model prediction, mesh-aligned evidences are extracted from finer-resolution features accordingly and fed back a regressor for parameter rectification. To enhance the reliability of spatial evidences, an auxiliary pixel-wise prediction task is imposed on the final output of the image encoder.}
		\vspace{-5mm}
		\label{fig:framework}
	\end{center}
\end{figure*}

\section{Methodology}\label{sec:methodology}

In this section, we will present the technical details of our approach.
As illustrated in Fig.~\ref{fig:framework}, our network produces a feature pyramid for mesh recovery in a coarse-to-fine fashion.
Coarse-aligned predictions will be improved by utilizing the mesh-aligned evidences extracted from spatial feature maps.
Moreover, an auxiliary prediction task is imposed on the image encoder, so that those spatial cues can be more reliable and relevant.

\subsection{Feature Pyramid for Human Mesh Regression}
The goal of our image encoder is to generate a pyramid of spatial features from coarse to fine granularities, which provide descriptions of the posed person in the image at different scale levels.
The feature pyramid will be used in subsequent predictions of the SMPL model with the pose, shape, and camera parameters $\Theta=\{\bm{\theta}, \bm{\beta}, \bm{\pi}\}$.

Formally, the encoder takes an image $I$ as input and outputs a set of spatial features $\{\bm{\phi}_s^t \in \mathbb{R}^{C_s\times H_s^t \times W_s^t}\}_{t=0}^{T-1}$ at the end, where $H_s^t$ and $W_s^t$ are monotonically increasing.
At level $t$, based on the feature map $\bm{\phi}_s^t$, a set of sampling points $X_t$ will be used to extract point-wise features.
Specifically, for each 2D point $x$ in $X_t$, point-wise features $\bm{\phi}_s^t(x) \in \mathbb{R}^{C_s\times 1}$ will be extracted from $\bm{\phi}_s^t$ accordingly using the bilinear sampling.
These point-wise features will go through a MLP (multi-layer perceptron) for dimension reduction and be further concatenated together as a feature vector $\bm{\phi}_p^t$, \ie,
\begin{equation}
    \bm{\phi}_p^t = \mathcal{F}(\bm{\phi}_s^t, X_t) = \oplus\left(\left\{f\left(\bm{\phi}_s^t(x)\right), \text{for}~x~\text{in}~X_t\right\}\right),
\label{ptfeat}
\end{equation}
where $\mathcal{F}(\cdot)$ denotes the feature sampling and processing operations, $\oplus$ denotes the concatenation, and $f(\cdot)$ is the MLP.
After that, a parameter regressor $\mathcal{R}_t$ takes features $\bm{\phi}_p^t$ and the current estimation of parameters $\Theta_t$ as inputs and outputs the parameter residual.
Parameters are then updated as $\Theta_{t+1}$ by adding the residual to $\Theta_t$.
For the level $t=0$, $\Theta_0$ adopts the mean parameters calculated from training data.

Given the parameter predictions $\Theta$ (the subscript $t$ is omitted for simplicity) at each level, a mesh with vertices of $M=\mathcal{M}(\bm{\theta}, \bm{\beta}) \in \mathbb{R}^{N\times 3}$ can be generated accordingly, where $N = 6890$ denotes the number of vertices in the SMPL model.
These mesh vertices can be mapped to sparse 3D joints $J\in \mathbb{R}^{N_j\times 3}$ by a pretrained linear regressor, and further projected on the image coordinate system as 2D keypoints $K=\bm{\Pi}(J)\in \mathbb{R}^{N_j\times 2}$, where $\bm{\Pi}(\cdot)$ denotes the projection function based on the camera parameters $\bm{\pi}$.
Note that the pose parameters in $\Theta$ are represented as relative rotations along kinematic chains and minor parameter errors can lead to large misalignment between the 2D projection and image evidences.
To penalize such misalignment during the training of the regression network, we follow common practices~\cite{kanazawa2018end,kolotouros2019learning} to add 2D supervisions on the 2D keypoints projected from the estimated mesh.
Meanwhile, additional 3D supervisions on 3D joints and model parameters are added when ground truth 3D labels are available.
Overall, the loss function for the parameter regressor is written as
\begin{equation}
\mathcal{L}_{reg} = \lambda_{2d}||K - \hat{K}|| + \lambda_{3d}||J - \hat{J}|| + \lambda_{para}||\Theta - \hat{\Theta}||,
\label{eq:loss_reg}
\end{equation}
where $||\cdot||$ is the squared L2 norm, $\hat{K}$, $\hat{J}$, and $\hat{\Theta}$ denote the ground truth 2D keypoints, 3D joints, and model parameters, respectively.

One of the improvements over the commonly used parameter regressors is that our regressors can better leverage spatial information.
Unlike the commonly used regressors taking the global features $\bm{\phi}_g \in \mathbb{R}^{C_g\times 1}$ as input, our regressor uses the point-wise information obtained from spatial features $\bm{\phi}_s^t$.
A straight-forward strategy to extract point-wise features would be using the points $X_t$ with a grid pattern and uniformly sampling features from $\bm{\phi}_s^t$.
In the proposed approach, the sampling points $X_t$ adopt the grid pattern at the level $t=0$ and will be updated according to the currently estimated mesh when $t>0$.
We will show that, such a mesh conditioned sampling strategy helps the regressor to produce more plausible results.

\subsection{Mesh Alignment Feedback Loop}

As mentioned in HMR~\cite{kanazawa2018end}, directly regressing mesh parameters in one go is challenging.
To tackle this issue, HMR uses an Iterative Error Feedback (IEF) loop to iteratively update $\Theta$ by taking the global features $\bm{\phi}_g$ and the current estimation of $\Theta$ as input.
Though the IEF strategy reduces parameter errors progressively, it uses the same global features each time for parameter update, which lacks fine-grained information and is not adaptive to new predictions.
By contrast, we propose a Mesh Alignment Feedback (MAF) loop so that mesh-aligned evidences can be leveraged in our regressor to rectify current parameters and improve the mesh-image alignment of the currently estimated model.

To this end, when $t>0$, we extract mesh-aligned features from $\bm{\phi}_s^t$ based on the currently estimated mesh $M_t$ to obtain more fine-grained and position-sensitive evidences.
Compared with the global features or the uniformly sampled grid features, mesh-aligned features can reflect the mesh-image alignment status of the current estimation, which is more informative for parameter rectification.
Specifically, the sampling points $X_t$ are obtained by first down-sampling the mesh $M_t$ to $\tilde{M}_t$ and then projecting it on the 2D image plane, \ie, $X_t = \bm{\Pi}(\tilde{M}_t)$.
Based on $X_t$, the mesh-aligned features $\bm{\phi}_m^t$ will be extracted from $\bm{\phi}_s^t$ using Eq.~\ref{ptfeat}, \ie, $\bm{\phi}_m^t = \bm{\phi}_p^t = \mathcal{F}(\bm{\phi}_s^t, \bm{\Pi}(\tilde{M}_t))$.
These mesh-aligned features will be fed into the regressor $\mathcal{R}_t$ for parameter update.
Overall, the proposed mesh alignment feedback loop can be formulated as
\begin{equation}
    \Theta_{t+1} = \Theta_t + \mathcal{R}_t\left(\Theta_t, \mathcal{F}(\bm{\phi}_s^t, \bm{\Pi}(\tilde{M}_t))\right), \text{for~} t > 0.
\end{equation}

\subsection{Auxiliary Pixel-wise Supervision}
\begin{figure}[t]
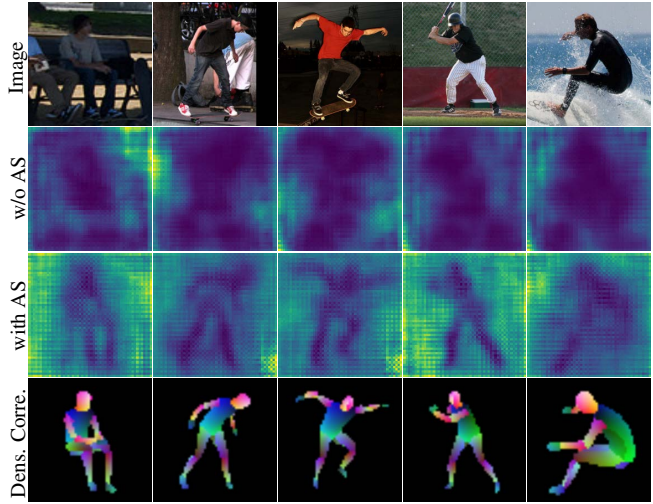

	\centering
% 	\hspace{1mm}
	\begin{tikzpicture}[remember picture,overlay]
	\node[font=\fontsize{8pt}{8pt}\selectfont, rotate=90] at (-4.2,-0.9) {Image};
	\node[font=\fontsize{8pt}{8pt}\selectfont, rotate=90] at (-4.2,-2.5) {w/o AS};
	\node[font=\fontsize{8pt}{8pt}\selectfont, rotate=90] at (-4.2,-4.2) {with AS};
	\node[font=\fontsize{8pt}{8pt}\selectfont, rotate=90] at (-4.2,-5.9) {Dens. Corre.};
	\end{tikzpicture}
% 	\hspace{0.5mm}
    \\
	\foreach \idx in {1,2,3,4,5} {
    	\begin{subfigure}[h]{0.085\textwidth}
    	    \includegraphics[width=1.1\textwidth]{fig/vis/fm/img_\idx.pdf}
        \end{subfigure}
	}
	\\
	\foreach \idx in {1,2,3,4,5} {
    	\begin{subfigure}[h]{0.085\textwidth}
    	    \includegraphics[width=1.1\textwidth]{fig/vis/fm/wo_dp_\idx.pdf}
        \end{subfigure}
	}
	\\
	\foreach \idx in {1,2,3,4,5} {
    	\begin{subfigure}[h]{0.085\textwidth}
    	    \includegraphics[width=1.1\textwidth]{fig/vis/fm/dp_\idx.pdf}
        \end{subfigure}
	}
	\\
	\foreach \idx in {1,2,3,4,5} {
    	\begin{subfigure}[h]{0.085\textwidth}
    	    \includegraphics[width=1.1\textwidth]{fig/vis/fm/iuv_\idx.pdf}
        \end{subfigure}
	}
	\\
	\vspace{-2mm}
	\caption{Visualization of the spatial feature maps and predicted dense correspondences. Top: Input images. Second / Third Row: Spatial feature maps learned without / with the Auxiliary Supervision (AS). Bottom: Predicted dense correspondence maps under the auxiliary supervision.}
	\vspace{-3mm}
	\label{fig:fm_auxsupv}
\end{figure}

As depicted in the second row of Fig.~\ref{fig:fm_auxsupv}, spatial features tend to be affected by the noisy inputs, since input images may contain a large amount of unrelated information such as occlusions, appearance and illumination variations.
To improve the reliability of the mesh-aligned cues extracted from spatial features, we impose an auxiliary pixel-wise prediction task on the spatial features at the last level.
Specifically, during training, the spatial feature maps $\bm{\phi}_s^{T-1}$ will go through a convolutional layer for the prediction of dense correspondence maps with pixel-wise supervision applied.
Dense correspondences encode the mapping relationship between foreground pixels on 2D image plane and mesh vertices in 3D space. In this way, the auxiliary supervision provides mesh-image correspondence guidance for the image encoder to preserve the most related information in the spatial feature maps.

In our implementation, we adopt the IUV maps defined in DensePose~\cite{alp2018densepose} as the dense correspondence representation, which consists of part index and UV values of the mesh vertices.
Note that we do not use DensePose annotations in the dataset but render IUV maps based on the ground-truth SMPL models~\cite{zhang2020learning}.
During training, classification and regression losses are applied on the part index $P$ and $UV$ channels of dense correspondence maps, respectively.
Specifically, for the part index $P$ channels, a cross-entropy loss is applied to classify a pixel belonging to either background or one among body parts.
For the $UV$ channels, a smooth L1 loss is applied to regress the corresponding $UV$ values of the foreground pixels.
Only the foreground regions are taken into account in the $UV$ regression loss, \ie, the estimated $UV$ channels are firstly masked by the ground-truth part index channels before applying the regression loss.
Overall, the loss function for the auxiliary pixel-wise supervision is written as
\begin{equation}
\begin{aligned}
\mathcal{L}_{aux} = &\lambda_{pi}CrossEntropy(P, \hat{P}) \\
            + &\lambda_{uv}SmoothL1(\hat{P} \odot U, \hat{P} \odot \hat{U}) \\
            + &\lambda_{uv}SmoothL1(\hat{P} \odot V, \hat{P} \odot \hat{V}),
\end{aligned}
\label{eq:aux_supv}
\end{equation}
where $\odot$ denotes the mask operation.
Note that the auxiliary prediction is required in the training phase only.

Fig.~\ref{fig:fm_auxsupv} visualizes the spatial features of the encoder trained with and without auxiliary supervision, where the feature maps are simply added along the channel dimension as grayscale images and visualized with colormap.
We can see that the spatial features are more neat and robust to the input variations when the auxiliary supervision is applied.

\section{Experiments}\label{sec:experiments}

% In this section, we present empirical results on tasks of 3D human pose and shape estimation as well as 2D segmentation and pose estimation.
% Moreover, ablation studies will be also conducted to validate the key components of the proposed PyMAF approach.

\subsection{Implementation Details}

The proposed PyMAF is validated on the ResNet-50~\cite{he2016deep} backbone pre-trained on ImageNet~\cite{deng2009imagenet}.
The ResNet-50 backbone takes a $224\times224$ image as input and produces image features with the size of $2048\times 7\times 7$.
For the classic regression network HMR~\cite{kanazawa2018end}, a $2048\times 1$ global feature vector will be obtained after average pooling.
In our approach, the image features will go through deconvolution layers, resulting spatial feature maps with resolutions of $\{14\times 14, 28\times 28, 56\times 56\}$, where $C_s=256$ for all resolutions. 
Here, the maximum number $T$ is set to $3$, which is equal to the iteration number used in HMR.
When generating mesh-aligned features, the SMPL mesh is down-sampled using a pre-computed downsampling matrix provided in~\cite{kolotouros2019convolutional}, after which the vertex number drops from $6890$ to $431$.
The mesh-aligned features of each point will be processed by a three-layer MLP so that their dimension will be reduced from $C_s$ to $5$.
Hence, the mesh-aligned feature vector has a length of $2155 = 431 \times 5$, which is similar to the length of global features.
For the grid features used at $t=0$, they are uniformly sampled from $\bm{\phi}_s^0$ with a $21\times21$ grid pattern, \ie, the point number is $441 = 21\times21$ which is approximate to the vertices number $431$ after mesh downsampling.
The regressors $\mathcal{R}_t$ have the same architecture with the regressor in HMR except that they have slightly different input dimensions.
Following the setting of SPIN~\cite{kolotouros2019learning}, we train our network using the Adam~\cite{kingma2014adam} optimizer with the learning rate set to $5e{-}5$ and the batch size of $64$ on a single 2080 Ti GPU.
No learning rate decay is applied during training.
More details of the implementation can be found in our code and the Supplementary Material.

\subsection{Datasets}
Following the settings of previous work~\cite{kanazawa2018end,kolotouros2019learning}, our approach is trained on a mixture of data from several datasets with 3D and 2D annotations, including Human3.6M~\cite{ionescu2014human3}, MPI-INF-3DHP~\cite{mehta2017monocular}, LSP~\cite{johnson2010clustered}, LSP-Extended~\cite{johnson2011learning}, MPII~\cite{andriluka20142d}, and COCO~\cite{lin2014microsoft}.
For the last five datasets, we also utilize their pseudo ground-truth SMPL parameters~\cite{bogo2016keep,kolotouros2019learning} for training.
We do not use training data from 3DPW~\cite{von2018recovering} but only perform evaluations on its test set.
Moreover, we do not use the DensePose annotations in COCO for auxiliary supervision, but render IUV maps based on the ground-truth SMPL meshes using the method described in~\cite{zhang2020learning}.
We evaluate our approach using a variety of metrics for quantitative comparisons with previous methods, \ie, PVE, MPJPE, and PA-MPJPE for the evaluation of 3D pose estimation, the segmentation accuracy, f1 scores, and AP for the measure of mesh-image alignment.
Detailed descriptions about the datasets and evaluation metrics can be found in the Supplementary Material.

\addtolength{\tabcolsep}{-5pt}
\begin{table}[t]
  \centering
  \footnotesize	
    \begin{tabular}{cl|ccc|cc}
    \toprule
    \multicolumn{2}{c|}{\multirow{2}[4]{*}{Method}} & \multicolumn{3}{c|}{3DPW} & \multicolumn{2}{c}{Human3.6M} \\
\cmidrule{3-7}    \multicolumn{2}{c|}{} & PVE   & MPJPE & PA-MPJPE & MPJPE & PA-MPJPE \\
    \midrule
    \multirow{5}[2]{*}{\begin{sideways}\rotatebox[origin=c]{0}{Temporal}\end{sideways}} & Kanazawa \etal~\cite{kanazawa2019learning} & 139.3 & 116.5 & 72.6 & -     & 56.9 \\
          & Doersch \etal~\cite{doersch2019sim2real} & -     & -     & 74.7  & -     & - \\
          & Arnab \etal~\cite{arnab2019exploiting} & -     & -     & 72.2 & 77.8  & 54.3 \\
          & DSD~\cite{sun2019human} & -     & -     & 69.5  & 59.1  & 42.4 \\
          & VIBE~\cite{kocabas2020vibe} & 113.4 & 93.5  & 56.5 & 65.9  & 41.5 \\
    \midrule
    \multirow{13}[4]{*}{\begin{sideways}\rotatebox[origin=c]{0}{Frame-based}\end{sideways}} & Pavlakos \etal~\cite{pavlakos2018learning} & -     & -     & -     & -     & 75.9 \\
          & HMR~\cite{kanazawa2018end} & -     & 130.0   & 76.7 & 88.0    & 56.8 \\
          & NBF~\cite{omran2018neural} & -     & -     &       & -     & 59.9 \\
          & GraphCMR~\cite{kolotouros2019convolutional} & -     & -     & 70.2 & -     & 50.1 \\
          & HoloPose~\cite{guler2019holopose} & -     & -     & -     & 60.3  & 46.5 \\
          & DenseRaC~\cite{xu2019denserac} & -     & -     & -     & 76.8  & 48.0 \\
          & SPIN~\cite{kolotouros2019learning} & 116.4 & 96.9  & 59.2 & 62.5  & 41.1 \\
          & DecoMR~\cite{zeng20203d} & -     & -     & 61.7$^\dagger$  & -     & \textbf{39.3}$^\dagger$ \\
          & DaNet~\cite{zhang2019danet} & -   &  -  &  56.9   & 61.5  & 48.6 \\
          & Song \etal~\cite{song2020human} & -     & -     &  \textbf{55.9}   & -  & 56.4 \\
          & I2L-MeshNet~\cite{moon2020i2l} & -     & 100.0   & 60.0    & \textbf{55.7}$^\dagger$  & 41.1$^\dagger$ \\
          & HKMR~\cite{georgakis2020hierarchical} & -     & -     & -     & 59.6  & 43.2 \\
\cmidrule{2-7}          & Baseline & 117.9 & 98.5  & 60.9  & 64.8  & 43.7 \\
          & PyMAF w/o AS & 113.6 & 95.6  & 58.8  & 60.3  & 42.3 \\
          & PyMAF & \textbf{110.1} & \textbf{92.8} & 58.9  & 57.7  & 40.5 \\
    \bottomrule
    \end{tabular}%
    \caption{Reconstruction errors on 3DPW and Human3.6M. $^\dagger$ denotes the numbers evaluated on non-parametric results.}
    \vspace{-5mm}
  \label{tab:3dpose}%
\end{table}%
\addtolength{\tabcolsep}{5pt}

\subsection{Comparison with the State of the Art}
% To demonstrate the superior performance of PyMAF, We evaluate our approach on benchmark datasets and make comparisons with the state of the art.

\textbf{3D Human Pose and Shape Estimation.}
We first evaluate our approach on the 3D human pose and shape estimation task, and make comparisons with previous state-of-the-art regression-based methods.
We present evaluation results for quantitative comparison on 3DPW and Human3.6M datasets in Table~\ref{tab:3dpose}.
Our PyMAF achieves competitive or superior results among previous approaches, including both frame-based and temporal approaches.
Note that the approaches reported in Table~\ref{tab:3dpose} are not strictly comparable since they may use different training data, learning rate schedules, or training epochs, \etc.
For a fair comparison, we report results of our baseline in Table~\ref{tab:3dpose}, which is trained under the same setting with PyMAF.
The baseline approach has the same network architecture with HMR~\cite{kanazawa2018end} and also adopts the 6D rotation representation~\cite{zhou2019continuity} for pose parameters.
Compared with the baseline, PyMAF reduces the MPJPE by 5.7 mm and 7.1 mm on 3DPW and Human3.6M datasets, respectively.
The auxiliary supervision (AS) also helps PyMAF to have better reconstruction results as shown in the last two rows of Table~\ref{tab:3dpose}.

From Table~\ref{tab:3dpose}, we can see that PyMAF has more notable improvements on the metrics MPJPE and PVE.
We would argue that the metric PA-MPJPE can not fully reveal the mesh-image alignment performance since it is calculated as the MPJPE after rigid alignment.
As depicted in the Supplementary Material, a reconstruction result with smaller PA-MPJPE value can have larger MPJPE value and worse alignment between the reprojected mesh and image.

\textbf{2D Segmentation and Pose Estimation.}
To quantitatively measure the mesh-image alignment of the predictions, we also conduct evaluation on the 2D segmentation and pose estimation task, where the predicted meshes are projected on the image plane to obtain 2D part segmentation and keypoints.
Table~\ref{tab:lsp} reports the assessment of foreground-background and six-part segmentation performance on the LSP test set.
As shown in Table~\ref{tab:lsp}, optimization-based approaches remain very competitive in terms of 2D alignment metrics and tend to outperform most of the regression-based ones.
The reason behind it is that optimization-based approaches is optimized for the mesh-image alignment explicitly.
Though PyMAF is regression-based, it surpasses all other methods including the optimization-based ones.
% Our closest competitor BodyNet~\cite{varol2018bodynet} adopts volumetric representations as its network output and needs post-processing to obtain parametric models, which is much more time-consuming.

\begin{table}[t]
\footnotesize
  \centering
    \begin{tabular}{lcccc}
    \toprule
    \multicolumn{1}{c}{\multirow{2}[4]{*}{Method}} & \multicolumn{2}{c}{FB Seg.} & \multicolumn{2}{c}{Part Seg.} \\
\cmidrule{2-5}          & acc.  & f1    & acc.  & f1 \\
    \midrule
    SMPLify oracle~\cite{bogo2016keep} & 92.17 & 0.88  & 88.82 & 0.67 \\
    SMPLify~\cite{bogo2016keep} & 91.89 & 0.88  & 87.71 & 0.64 \\
    SMPLify on~\cite{pavlakos2018learning} & 92.17 & 0.88  & 88.24 & 0.64 \\
    \midrule
    HMR~\cite{kanazawa2018end} & 91.67 & 0.87  & 87.12 & 0.60 \\
    BodyNet~\cite{varol2018bodynet} & 92.75 & 0.84  & -     & - \\
    CMR~\cite{kolotouros2019convolutional} & 91.46 & 0.87  & 88.69 & 0.66 \\
    TexturePose~\cite{pavlakos2019texturepose} & 91.82 & 0.87  & 89.00 & 0.67 \\
    SPIN~\cite{kolotouros2019learning} & 91.83 & 0.87  & 89.41 & 0.68 \\
    DecoMR~\cite{zeng20203d} & 92.10 & 0.88  & 89.45 & 0.69 \\
    HKMR~\cite{georgakis2020hierarchical} & 92.23 & 0.88  & 89.59 & 0.69 \\
    \midrule
    Baseline & 91.67 & 0.87  & 89.23 & 0.68 \\
    PyMAF w/o AS & 92.43 & 0.88  & 89.98 & 0.70 \\
    PyMAF & \textbf{92.79} & \textbf{0.89} & \textbf{90.47} & \textbf{0.72} \\
    \bottomrule
    \end{tabular}%
    \caption{Foreground-background and six-part segmentation accuracy and f1 scores on the LSP test set. SMPLify Oracle denotes the SMPLify using ground-truth keypoints as inputs.}
    \vspace{-5mm}
  \label{tab:lsp}%
\end{table}%

\begin{figure}[t]
	\centering
% 	\hspace{1mm}
	\begin{tikzpicture}[remember picture,overlay]
	\node[font=\fontsize{8pt}{8pt}\selectfont, rotate=90] at (0,2.7) {Image};
 	\node[font=\fontsize{8pt}{8pt}\selectfont, rotate=90] at (0,1.1) {SPIN~\cite{kolotouros2019learning}};
	\node[font=\fontsize{8pt}{8pt}\selectfont, rotate=90] at (0,-0.5) {Baseline};
	\node[font=\fontsize{8pt}{8pt}\selectfont, rotate=90] at (0,-2.1) {PyMAF};
	\end{tikzpicture}
% 	\hspace{0.5mm}
	\foreach \idx in {1,2,3,4,5} {
		\begin{subfigure}[h]{0.082\textwidth}
			\centering
			\foreach \sub in {0,1,2,3} {
    			\pgfmathsetmacro\imidx{int(\sub * 5 + \idx)}
    			\includegraphics[width=1.1\textwidth]{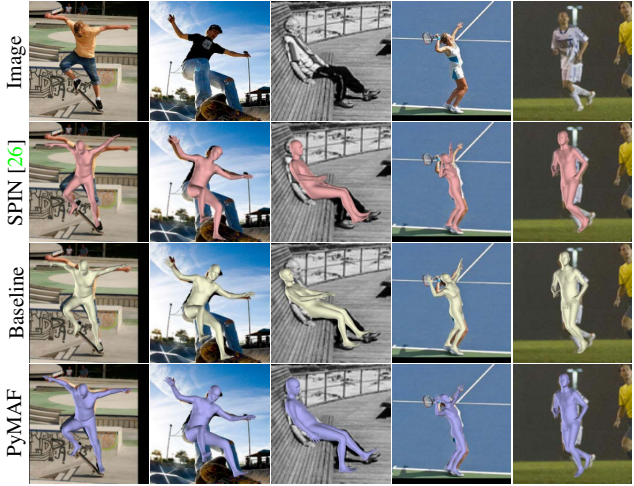}
    % 			\vspace{1mm}
    		}
		\end{subfigure}
	}
	\vspace{-6mm}
	\caption{Qualitative comparison of reconstruction results on the COCO validation set.}
	\vspace{-2mm}
	\label{fig:cocoDemo}
\end{figure}

Finally, we evaluate 2D human pose estimation performance on the COCO validation set to verify the effectiveness of our approach in real-world scenarios.
During the evaluation, we project keypoints from the estimated mesh on the image plane, and compute the Average Precision (AP) based on the keypoint similarity with the ground truth 2D keypoints.
The results of keypoint localization APs are reported in Table~\ref{tab:coco}.
OpenPose~\cite{cao2019openpose}, a widely-used 2D human pose estimation algorithm, are also included for reference.
We can see that, the COCO dataset is very challenging for approaches to human mesh reconstruction as they typically have much worse performances in terms of 2D keypoint localization accuracy.
In Table~\ref{tab:coco}, we also include the results of the optimization-based SMPLify~\cite{bogo2016keep} by fitting the SMPL model to the ground-truth 2D keypoints.
As pointed out in previous work~\cite{kolotouros2019learning}, SMPLify may produce well-aligned but unnatural results.
Moreover, SMPLify is much more time-consuming than regression-based solutions.
Among approaches recovering 3D human mesh, PyMAF outperforms previous regression-based methods by remarkable margins.
Compared with our baseline, PyMAF significantly improves the $\textrm{AP}$ and $\textrm{AP}_{50}$ by 7.8\% and 10.7\%, respectively.
The auxiliary supervision (AS) also considerably contributes to more robust reconstruction in this challenging dataset and brings 3.9\% performance gain on $\textrm{AP}$.
Reconstruction results on COCO are depicted in Fig.~\ref{fig:cocoDemo} for qualitative comparisons, where PyMAF convincingly performs better than SPIN~\cite{kolotouros2019learning} and our baseline by producing better-aligned and natural results.

\begin{table}[t]
 \footnotesize
  \centering
    \begin{tabular}{l|ccccc}
    \toprule
    Method & AP    & $\textrm{AP}_{50}$ & $\textrm{AP}_{75}$ & $\textrm{AP}_{M}$ & $\textrm{AP}_{L}$ \\
    \midrule
    OpenPose~\cite{cao2019openpose} & \textbf{65.3}  & \textbf{85.2}  & \textbf{71.3}  & \textbf{62.2}  & \textbf{70.7} \\
    \midrule
    SMPLify~\cite{bogo2016keep} & 22.0  & 37.7  & \textbf{23.1}  & \textbf{27.7}  & 17.6 \\
    HMR~\cite{kanazawa2018end} & 18.9  & 47.5  & 11.7  & 21.5  & 17.0 \\
    GraphCMR~\cite{kolotouros2019convolutional} & 9.3   & 26.9  & 4.2   & 11.3  & 8.1 \\
    SPIN~\cite{kolotouros2019learning} & 17.3  & 39.1  & 13.5  & 19.0    & 16.6 \\
    \midrule
    Baseline & 16.8  & 38.2  & 12.8  & 18.5  & 16.0   \\
    PyMAF w/o AS & 20.7  & 43.9  & 17.4  & 22.3  & 19.9  \\
    PyMAF & \textbf{24.6} & \textbf{48.9} & 22.7 & 26.0 & \textbf{24.2} \\
    \bottomrule
    \end{tabular}%
    \caption{Keypoint localization APs on the COCO validation set. There is a total of 50,197 samples used for evaluation. Results of SMPLify~\cite{bogo2016keep} are evaluated based on the implementation in SPIN~\cite{kolotouros2019learning} with 300 optimization iterations. Results of HMR~\cite{kanazawa2018end}, GraphCMR~\cite{kolotouros2019convolutional}, and SPIN~\cite{kolotouros2019learning} are evaluated based on their publicly released code and models.}
    \vspace{-5mm}
  \label{tab:coco}%
\end{table}%

\subsection{Ablation Study}
In this part, we will perform ablation studies under various settings on Human3.6M to validate the efficacy of the key components proposed in our approach.
All ablation approaches are trained and tested on Human3.6M, as it includes ground-truth 3D labels and is the most widely used benchmark for 3D human pose and shape estimation.

\textbf{Efficacy of Mesh-aligned Features.}
In PyMAF, mesh-aligned features provide the current mesh-image alignment information in the feedback loop, which is essential for better mesh recovery.
To verify this, we alternatively replace mesh-aligned features with the global features or the grid features uniformly sampled from spatial features as the input for parameter regressors.
Table~\ref{tab:MeshAligned} reports the performances of the approaches equipped with different types of features in the feedback loop.
The results under the non-pyramidal setting are also included in Table~\ref{tab:MeshAligned}, where the grid and mesh-aligned features are extracted from the feature maps with the highest resolution (\ie, $56 \times 56$) and the mesh-aligned features are extracted on the reprojected points of the mesh under the mean pose at $t=0$.
For fair comparisons with Baseline, the approaches with global, grid, and mesh-aligned feedback features under the non-pyramidal setting also use a single regressor but have individual supervision on the prediction of each iteration.
Besides, all approaches in Table~\ref{tab:MeshAligned} do not use the auxiliary supervision.

Unsurprisingly, using mesh-aligned features yields the best performance under both non-pyramidal and pyramidal designs.
The approach using the grid features sampled from spatial feature maps has better results than that using the global features but is worse than the mesh-aligned counterpart.
When using pyramidal feature maps, the mesh-aligned solution achieves even more performance gain since multi-scale mesh-alignment evidences can be leveraged in the feedback loop.
Though the grid features largely contain spatial cues on uniformly distributed pixel positions, they can not reflect the alignment status of the current estimation.
This implies that mesh-aligned features are the most informative one for the regressor to rectify the current mesh parameters.

\begin{table}[t]
  \centering
  \footnotesize
    \begin{tabular}{l|c|c|cc}
    \toprule
    Feedback Feat. & Pyramid? & \# Regressor & MPJPE & PA-MPJPE \\
    \midrule
    Global (Baseline) & No    & 1     & 84.1  & 55.6 \\
    \midrule
    Global & \multirow{3}[2]{*}{No} & \multirow{3}[2]{*}{1} & 84.3  & 55.3 \\
    Grid  &       &       & 80.5  & 54.7 \\
    Mesh-aligned &       &       & \textbf{79.6} & \textbf{53.4} \\
    \midrule
    Grid  & \multirow{2}[2]{*}{Yes} & \multirow{2}[2]{*}{3} & 79.7  & 54.3 \\
    Mesh-aligned &       &       & \textbf{76.8} & \textbf{50.9} \\
    \bottomrule
    \end{tabular}%
    \caption{Ablation study on using different types of feedback features for refinement. No auxiliary supervision is applied.}
    \vspace{-5mm}
  \label{tab:MeshAligned}%
\end{table}%

\textbf{Benefit from Auxiliary Supervision.}
The auxiliary pixel-wise supervision helps to enhance the reliability of the mesh-aligned evidences extracted from spatial features.
Using alternative pixel-wise supervision such part segmentation rather than dense correspondences is also possible in our framework.
In our approach, these auxiliary predictions are solely needed for supervisions during training since the point-wise features are extracted from feature maps.
For more in-depth analyses, we have also tried extracting point-wise features from the auxiliary predictions, \ie, the input type of regressors are intermediate representations such as part segmentation or dense correspondences.
Table~\ref{tab:AuxSupv} shows the comparison of experiments with different auxiliary supervision settings and input types for regressors during training.
Using part segmentation is slightly worse than our dense correspondence solution.
Compared with the part segmentation, the dense correspondences preserve not only clean but also rich information in foreground regions.
Moreover, using feature maps for point-wise feature extraction consistently performs better than using auxiliary predictions.
This can be explained by the fact that using intermediate representations as input for regressors hampers the end-to-end learning of the whole network.
Under the auxiliary supervision strategy, the spatial feature maps are learned with the signal backpropagated from both auxiliary prediction and parameter correction tasks.
In this way, the background features can also contain information for mesh parameter correction since the deep features have larger receptive fields and are trained in an end-to-end manner.
% Moreover, the mesh-aligned features are designed to reflect the misalignment status of the current mesh estimation and hence need to contain both foreground and background features.
As shown in Table~\ref{tab:AuxSupv}, when the mesh-aligned features are masked with the foreground region of part segmentation predictions, the performance degrades from 75.5 mm to 77.6 mm on MPJPE.

\begin{table}[t]
\footnotesize
  \centering
    \begin{tabular}{c|c|cc}
    \toprule
    Aux. Supv. & Input Type & MPJPE & PA-MPJPE \\
    \midrule
    None  & Feature & 76.8  & 50.9 \\
    \midrule
    \multirow{3}[2]{*}{Part. Seg.} & Part. Seg. & 108.1 & 75.9 \\
          & Feature & 75.5  & 49.2 \\
          & Feature*Part. Seg. & 77.6  & 51.1 \\
    \midrule
    \multirow{2}[2]{*}{Dense Corr.} & Dense Corr. & 77.8  & 54.7 \\
          & Feature & \textbf{75.1} & \textbf{48.9} \\
    \bottomrule
    \end{tabular}%
    \caption{Ablation study on using different auxiliary supervision settings and input types for regressors.}
    \vspace{-6mm}
  \label{tab:AuxSupv}%
\end{table}%

\begin{figure}[t]
	\centering
% 	\hspace{1mm}
% 	\hspace{0.5mm}
	\foreach \idx in {1,2,3,4,5} {
		\begin{subfigure}[h]{0.45\textwidth}
% 			\centering
    		\includegraphics[width=1\textwidth]{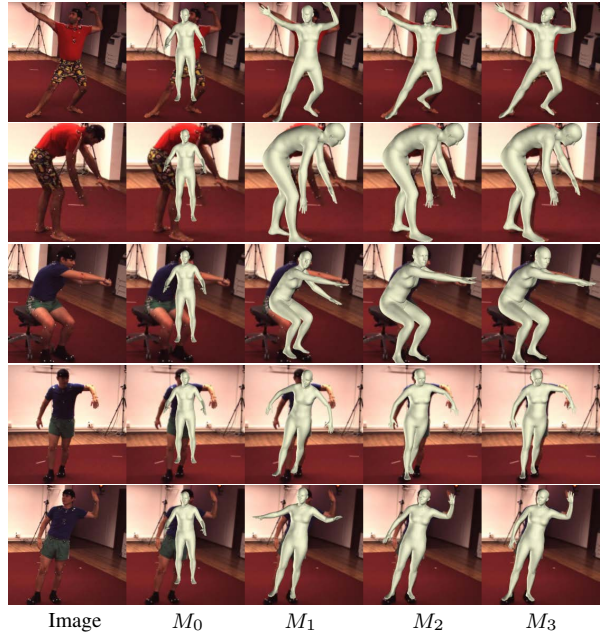}
		\end{subfigure}
	}
	\\
	\begin{tikzpicture}[remember picture,overlay]
	\node[font=\fontsize{8pt}{8pt}\selectfont] at (-3,-0.2) {Image};
	\node[font=\fontsize{8pt}{8pt}\selectfont] at (-1.5,-0.2) {$M_0$};
 	\node[font=\fontsize{8pt}{8pt}\selectfont] at (-0,-0.2) {$M_1$};
	\node[font=\fontsize{8pt}{8pt}\selectfont] at (1.7,-0.2) {$M_2$};
	\node[font=\fontsize{8pt}{8pt}\selectfont] at (3.2,-0.2) {$M_3$};
	\end{tikzpicture}
	\vspace{1mm}
	\caption{Visualization of reconstruction results across different iterations in the feedback loop.}
	\vspace{-5mm}
	\label{fig:loop}
\end{figure}

\textbf{Effect of the Initialization in the Feedback Loop.}
In our approach, the point-wise features are initially extracted on grid points for coarse mesh predictions before the extraction of mesh-aligned features.
It is also feasible to extract mesh-aligned features based on a mean-pose mesh at $t=0$ as the initial features.
The performances of the approaches using different initial features are reported in Table~\ref{tab:iter}, where PyMAF can improve reconstruction results under both initialization settings.
For the approach using the initial features extracted on the projection of the mean-pose mesh, we visualize its estimated meshes after each iteration in Fig~\ref{fig:loop}.
Though the mean-pose mesh is far away from the ground-truth, PyMAF can correct the drift of body parts progressively and result in better-aligned human models.

\begin{table}[t]
\footnotesize
  \centering
    \begin{tabular}{c|l|cccc}
    \toprule
    Initial Feat. & \multicolumn{1}{c|}{Metric} & $M_0$ & $M_1$ & $M_2$ & $M_3$ \\
    \midrule
    Mean-pose & MPJPE & 274.0 & 81.4  & 78.0  & 77.3 \\
    Mesh  & PA-MPJPE & 131.7 & 54.1  & 51.1  & 50.3 \\
    \midrule
    \multirow{2}[2]{*}{Grid Points} & MPJPE & 274.0 & 80.3  & 76.6  & 75.1 \\
          & PA-MPJPE & 131.7 & 52.1  & 49.9  & 48.9 \\
    \bottomrule
    \end{tabular}%
    \vspace{-1mm}
    \caption{Ablation study on using different initial features across iterations in the feedback loop.}
    \vspace{-2mm}
  \label{tab:iter}%
\end{table}%

%------------------------------------------------------------------------
% \section{Conclusion}

% In this paper, we present Pyramidal Mesh Alignment Feedback (PyMAF) for regression-based human mesh recovery.
% At the core of our approach, the parameter regressor leverages spatial information from a feature pyramid, and the parameter deviation can be corrected explicitly in a feedback loop based on the alignment status of the currently estimated meshes.
% To achieve this, given coarse-aligned mesh predictions, the mesh-aligned features are first extracted from spatial feature maps and then fed back into the regressor for parameter rectification.
% Moreover, an auxiliary dense correspondence task is imposed on the spatial feature maps during the training procedure of our deep regressor.
% In this way, the pixel-wise supervision signals provide guidance to enhance the relevance and reliability of the mesh-aligned features.
% The efficacy of PyMAF is validated on both indoor and in-the-wild datasets, where the proposed approach can consistently improve the mesh-image alignment performance over the baseline and previous regression-based approaches.
\section{Limitations and Future Work}

In this paper, we present Pyramidal Mesh Alignment Feedback (PyMAF) for regression-based human mesh recovery.
PyMAF is primarily motivated by the observation of the reprojection misalignment of parametric mesh results.
Though PyMAF improves the mesh-image alignment on 2D image planes, it can still hardly address the depth ambiguity problem in 3D space.
Moreover, PyMAF fails to handle extreme shapes due to the lack of training data, as shown in the Supplementary Material.

In the future, PyMAF can be extended and incorporated with the recent progress to improve the alignment in 3D space~\cite{luan2021pc,li2020hybrik}.
Moreover, combining PyMAF with HoloPose~\cite{guler2019holopose}, SPIN~\cite{kolotouros2019learning}, or EFT~\cite{joo2020exemplar} for the generation of more precise pseudo 3D ground-truth labels would also be interesting future work.

\small{\paragraph{\small \bf Acknowledgements.} This work was supported by the National Natural Science Foundation of China (Grant U1836217, 62125107, 62076119, 61921006), Australian Research Council (DP200103223), and Australian Medical Research Future Fund (MRFAI000085). Wanli Ouyang is also supported by SenseTime.}

{\small
\bibliographystyle{ieee_fullname}
\bibliography{egbib}
}

\newpage

\appendix

\section{Appendix}

\subsection{More Experimental Details}
\label{sec:implementation}
Our network is trained with the Adam~\cite{kingma2014adam} optimizer and batch size of 64.
The learning rate is set to $5e{-}5$ without learning rate decay during training.
Similar to SPIN~\cite{kolotouros2019learning}, our network is first trained on Human3.6M for 60 epochs and then on the mixture of both 2D and 3D datasets for another 60 epochs.

The parameter regressors of PyMAF have the same design with that of HMR~\cite{kanazawa2018end} except for their slightly different input and output dimensions.
Specifically, a regressor consists of two fully-connected layers each with 1024 hidden neurons and dropout added in between, followed by a final layer at the end with 157-dimension output, corresponding to the residual of shape and pose parameters.
The regressors in our network adopt the continuous representation~\cite{zhou2019continuity} for 3D rotations in the pose parameters $\bm{\theta}$.
During the extraction of mesh-aligned features, the dimension of point-wise features is reduced from $C_s$ (\ie, 256) to 5, where a three-layer MLP consisting of two hidden layers with neuron numbers of (128, 64) is used.
The feature pyramid of PyMAF is generated by three deconvolution layers. The deconvolutions are not compulsory but help to produce better features maps.
In our experiments, using the feature maps in the earlier layers is also feasible but inferior to our final solution.

\textbf{Runtime.} The PyTorch implementation of PyMAF takes about 30 ms to process one sample on the machine with a single 2080 Ti GPU.
The proposed mesh alignment feedback loop takes about 6 ms for each iteration, including the time of generating new feature maps via deconvolution, projecting the mesh on image planes, the extraction of mesh-aligned features via bilinear sampling and MLPs, and the prediction of parameter updates by the regressor.
For each iteration, compared to the feedback loop in HMR~\cite{kanazawa2018end} or SPIN~\cite{kolotouros2019learning}, PyMAF introduces additional runtime in the generation of feature maps, the current SMPL meshes, and the mesh-aligned features, which accounts for 0.3 ms, 4 ms, and 1.2 ms respectively.
We can see that the generation of the feature pyramid and mesh-aligned features is quite efficient, and the main runtime overhead comes from the SMPL mesh generation given the current parameters.
In practice, we can speed up this process by using a down-sampled version of SMPL to generate the mesh with 431 vertices directly.
Note that the prediction of dense correspondences and the auxiliary supervision in the pipeline are needed for training only, which accounts for additional 15\% runtime.

\subsection{Datasets}
\label{sec:datasets}
Following the protocols of previous work~\cite{kanazawa2018end,kolotouros2019learning}, we train our network on several datasets with 3D or 2D annotations, including Human3.6M~\cite{ionescu2014human3}, MPI-INF-3DHP~\cite{mehta2017monocular}, LSP~\cite{johnson2010clustered}, LSP-Extended~\cite{johnson2011learning}, MPII~\cite{andriluka20142d}, COCO~\cite{lin2014microsoft}.
For the last five datasets, we also incorporate the SMPL parameters fitted in \cite{bogo2016keep,kolotouros2019learning} as pseudo groud-truth annotations for training.
Here, we provide more descriptions of the datasets to supplement the main manuscript.

\textbf{3DPW}~\cite{von2018recovering} is captured in challenging outdoor scenes with IMU-equipped actors under various activities.
This dataset provides accurate shape and pose ground truth annotations.
Following the protocol of previous work~\cite{kanazawa2019learning,kolotouros2019learning}, we do not use its data for training but only perform evaluations on its test set.

\textbf{Human3.6M}~\cite{ionescu2014human3} is commonly used as the benchmark dataset for 3D human pose estimation, consisting of 3.6 million video frames captured in the controlled environment.
The ground truth SMPL parameters in Human3.6M are generated by applying MoSh~\cite{loper2014mosh} to the sparse 3D MoCap marker data, as done in Kanazawa \etal~\cite{kanazawa2018end}.
It is common protocols~\cite{pavlakos2017coarse,pavlakos2018learning,kanazawa2018end} to use five subjects (S1, S5, S6, S7, S8) for training and two subjects (S9, S11) for evaluation.
The original videos are also down-sampled from 50 fps to 10 fps to remove redundant frames, resulting in 312,188 frames for training and 26,859 frames for evaluation.

\textbf{MPI-INF-3DHP}~\cite{mehta2017monocular} is a recently introduced 3D human pose dataset covering more actor subjects and poses than Human3.6M.
The images of this dataset were collected under both indoor and outdoor scenes, and the 3D annotations were captured by a multi-camera marker-less MoCap system.
Hence, there are some noise in the 3D ground truth annotations.

\textbf{LSP-Extended}~\cite{johnson2011learning} is a 2D human pose benchmark dataset, containing person images with challenging poses. There are 14 visible 2D keypoint locations annotated for each image and 9,428 samples used for training.

\textbf{LSP}~\cite{johnson2010clustered} is a standard benchmark dataset for 2D human pose estimation. In our experiments, we will employ its test set for silhouette/parts segmentation evaluation, where the annotations come from Lassner \etal~\cite{lassner2017unite}.
There are 1,000 samples used for evaluation.

\textbf{MPII}~\cite{andriluka20142d} is a standard benchmark for 2D human pose estimation.
There are 25,000 images collected from YouTube videos covering a wide range of activities.
We discard those images without complete keypoint annotations, producing 14,810 samples for training.

\textbf{COCO}~\cite{lin2014microsoft} contains a large scale of person images labeled with 17 keypoints.
In our experiments, we only use those persons annotated with at least 12 keypoints, resulting in 28,344 samples for training.
Since this dataset do not contain ground-truth meshes, we conduct quantitative evaluation on the 2D keypoint localization task using its validation set, which consists of 50,197 samples.
Following~\cite{zhang2020learning}, we crop input images using the ground-truth bounding boxes.

\begin{figure}[ht]
	\centering
	\begin{subfigure}[b]{0.23\textwidth}
		\includegraphics[width=1\textwidth]{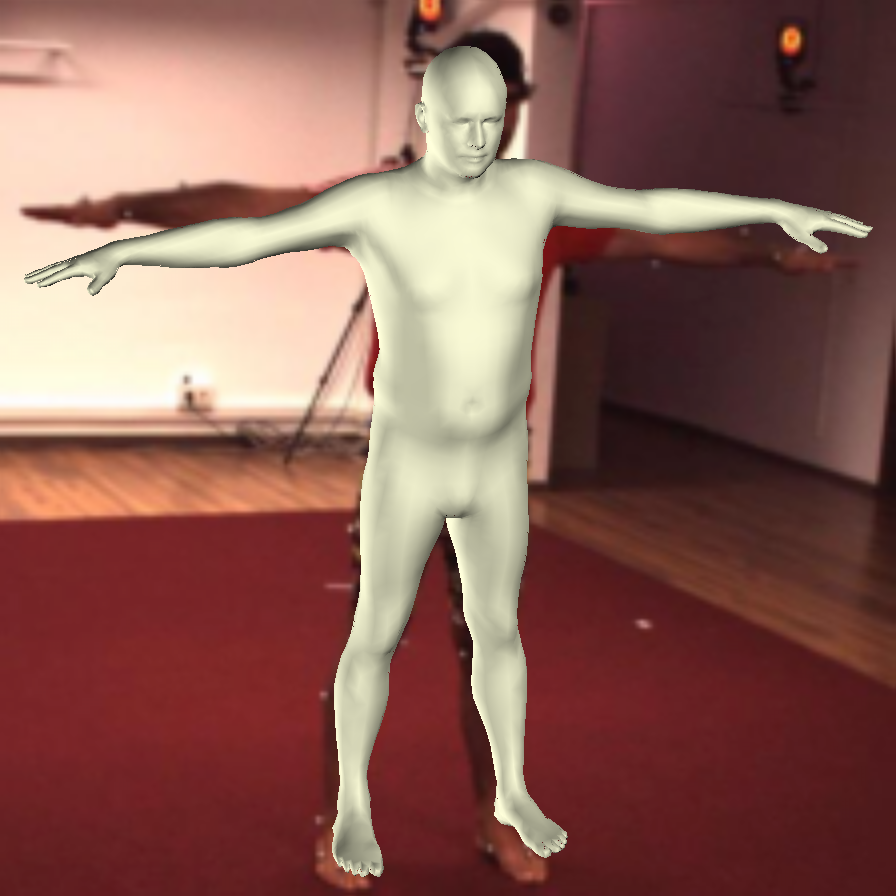}
		\caption{\scriptsize PA-MPJPE: 26.9, MPJPE: 74.3}
		\label{fig:bad}
    \end{subfigure}
	\begin{subfigure}[b]{0.23\textwidth}
		\includegraphics[width=1\textwidth]{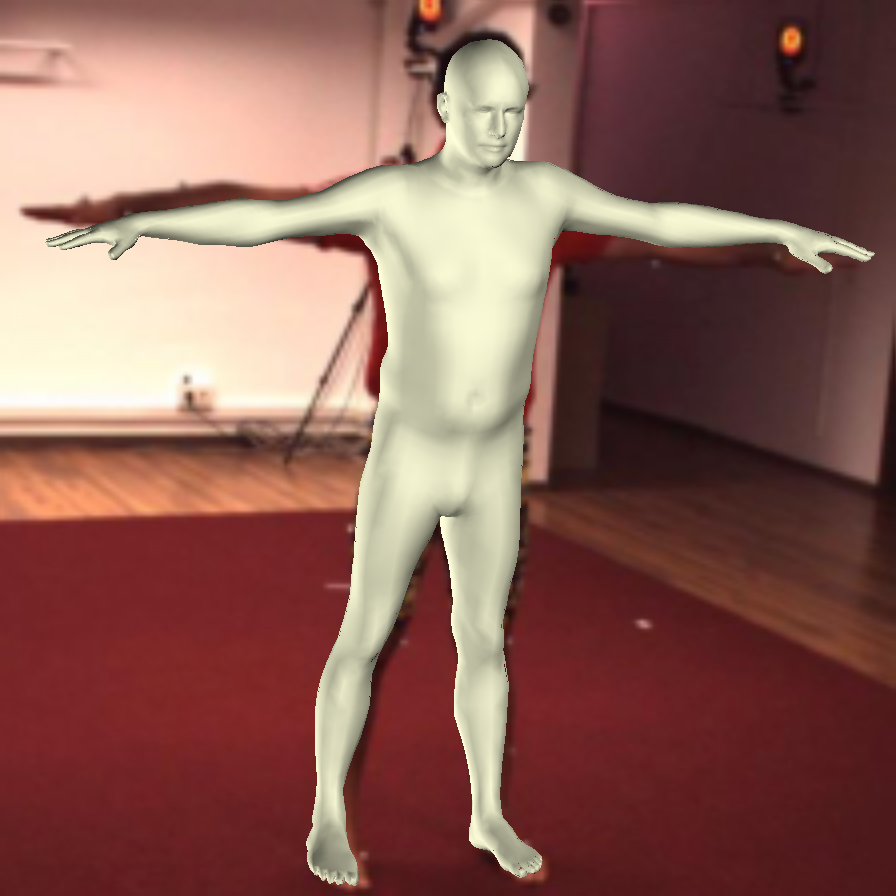}
		\caption{\scriptsize PA-MPJPE: 27.7, MPJPE: 43.4}
		\label{fig:good}
    \end{subfigure}
    \vspace{-2mm}
	\caption{Examples of two reconstruction results. (a) A reconstruction result with a lower PA-MPJPE value but worse mesh-image alignment. (b) A reconstruction result with a higher PA-MPJPE value but better mesh-image alignment.}
	\label{fig:metrics}
\end{figure}

\subsection{Evaluation Metrics}
\label{sec:metric}
In the main manuscript, we report results of our approach in a variety of evaluation metrics for quantitative comparisons with the state of the art, where all metrics are computed in the same way as previous work~\cite{kanazawa2018end,pavlakos2018learning,kolotouros2019learning} in the literature.

To quantitatively evaluate the 3D human reconstruction and pose estimation performance on 3DPW and Human3.6M, PVE, MPJPE, and PA-MPJPE are adopted as the evaluation metrics in Table~\ref{tab:3dpose} of the main manuscript.
They are all reported in millimeters (mm) by default.
Among these three metrics, PVE denotes the mean Per-vertex Error defined as the average point-to-point Euclidean distance between the predicted mesh vertices and the ground truth mesh vertices.
MPJPE denotes the Mean Per Joint Position Error, and PA-MPJPE denotes MPJPE after rigid alignment of the prediction with ground truth using Procrustes Analysis.
Note that the metric PA-MPJPE can not fully reveal the mesh-image alignment performance since it is calculated as MPJPE after rigid alignment.
As depicted in Fig.~\ref{fig:metrics}, a reconstruction result with a lower PA-MPJPE value can have a higher MPJPE value and worse alignment between the reprojected mesh and image.

In Table~\ref{tab:lsp} of the main manuscript, segmentation accuracy metrics quantitatively measure the mesh-image alignment of different approaches on the LSP dataset.
As originally done in Lassner \etal~\cite{lassner2017unite}, silhouette (\ie, Foreground/Background, FB) and Part segmentation are considered in calculating the accuracy and f1 scores.

For 2D human pose estimation task on COCO\footnote{\url{https://cocodataset.org/\#keypoints-eval}}, the commonly-used Average Precision (AP) is adopted as the evaluation metric.
AP is calculated based on the Object Keypoint Similarity (OKS), which plays a similar role as IoU in object detection.
In Table~\ref{tab:coco} of the main manuscript, the results are reported using mean AP, and variants of AP including $\textrm{AP}_{50}$ (AP at OKS = 0.50), $\textrm{AP}_{75}$ (AP at OKS = 0.75), $\textrm{AP}_{M}$ for persons with medium sizes, and $\textrm{AP}_{L}$ for persons with large sizes.

\subsection{More Experimental Results}
\label{sec:quan_eval}

\begin{figure*}[t]
	\centering
% 	\hspace{1mm}
% 	\hspace{0.5mm}
	\foreach \idx in {1, 2, 3, 5, 4, 6} {
	    \includegraphics[width=0.48\textwidth]{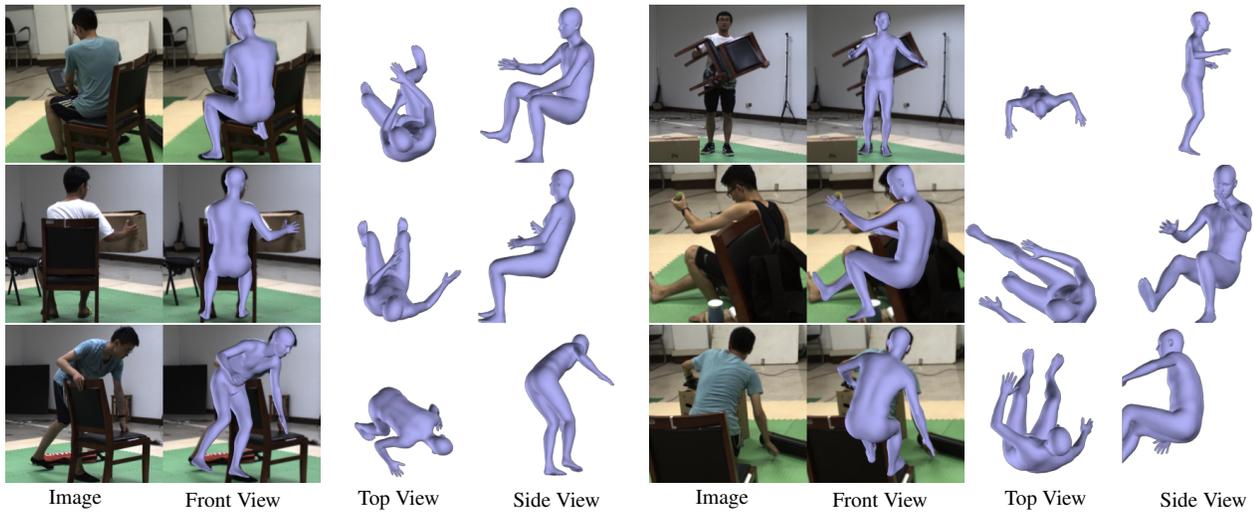}
	}\\
	\begin{tikzpicture}[remember picture,overlay]
	\node[font=\fontsize{8pt}{8pt}\selectfont] at (-7.5,0.2) {Image};
	\node[font=\fontsize{8pt}{8pt}\selectfont] at (-5.4,0.2) {Front View};
 	\node[font=\fontsize{8pt}{8pt}\selectfont] at (-3.2,0.2) {Top View};
	\node[font=\fontsize{8pt}{8pt}\selectfont] at (-1.1,0.2) {Side View};
	\node[font=\fontsize{8pt}{8pt}\selectfont] at (1.1,0.2) {Image};
	\node[font=\fontsize{8pt}{8pt}\selectfont] at (3.2,0.2) {Front View};
 	\node[font=\fontsize{8pt}{8pt}\selectfont] at (5.4,0.2) {Top View};
	\node[font=\fontsize{8pt}{8pt}\selectfont] at (7.5,0.2) {Side View};
	\end{tikzpicture}
	\vspace{-2mm}
	\caption{Reconstruction results of PyMAF on the 3DOH50K~\cite{zhang2020object} dataset. PyMAF helps to handle occlusions.}
	\label{fig:3doh50k}
\end{figure*}

\begin{figure*}[t]
	\centering
% 	\hspace{1mm}
% 	\hspace{0.5mm}
	\foreach \idx in {1, 2} {
	    \includegraphics[width=0.48\textwidth]{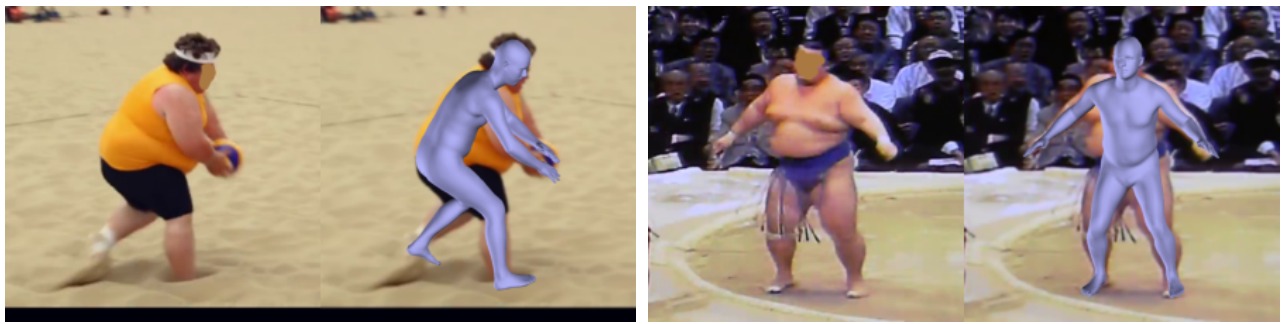}
	}
	\vspace{-2mm}
	\caption{Reconstruction results of PyMAF on the SSP-3D~\cite{sengupta2020synthetic} dataset. PyMAF fails to handle extreme shapes due to the lack of training data.}
	\label{fig:ssp}
\end{figure*}

\textbf{More Quantitative Results.} To evaluate the performances of PyMAF on human images with occlusions and different body shape styles, we conduct evaluation experiments on 3DOH50K~\cite{zhang2020object} and SSP-3D~\cite{sengupta2020synthetic} datasets.
The test set of 3DOH50K includes 1,290 person images in occlusion scenarios, while SSP-3D consists of 311 images of sport persons with a variety of body shapes and poses.
Note that we only perform testing on these two datasets and do not use their data for training.
Experimental results on 3DOH50K and SSP-3D are reported in Tab.~\ref{tab:3doh_ssp}, and
qualitative results are shown in Figures~\ref{fig:3doh50k} and~\ref{fig:ssp}.
PyMAF can improve the reconstruction under occlusions on 3DOH50K and help with more accurate shape estimation on SSP-3D.
Despite the numerical performance gains, PyMAF fails to handle extreme shapes on the SSP-3D dataset, as shown in Figure~\ref{fig:ssp}.

\begin{table}[htbp]
  \centering
    \begin{tabular}{l|cc|c}
    \toprule
          & \multicolumn{2}{c|}{3DOH50K} & \multicolumn{1}{l}{SSP-3D} \\
    \midrule
          & PVE$\downarrow$ & MPJPE$\downarrow$ & mIOU$\uparrow$ \\
    \midrule
    SPIN~\cite{kolotouros2019learning}  & 113.4 & 102.3 & 70.2 \\
    Baseline & 113.1 & 102.0   & 70.8 \\
    PyMAF & \textbf{107.3} & \textbf{96.2}  & \textbf{72.1} \\
    \bottomrule
    \end{tabular}%
    \caption{Reconstruction performances on 3DOH50K and SSP-3D datasets.}
  \label{tab:3doh_ssp}%
\end{table}%

\begin{figure*}[t]
	\centering
% 	\hspace{1mm}
% 	\hspace{0.5mm}
	\begin{tikzpicture}[remember picture,overlay]
 	\node[font=\fontsize{8pt}{8pt}\selectfont, rotate=90] at (-8.7,-1.1) {SPIN~\cite{kolotouros2019learning}};
	\node[font=\fontsize{8pt}{8pt}\selectfont, rotate=90] at (-8.7,-3.1) {PyMAF};
	\end{tikzpicture}
	\\
	\foreach \idx in {1, 2, 3, 4} {
	    \includegraphics[width=0.48\textwidth]{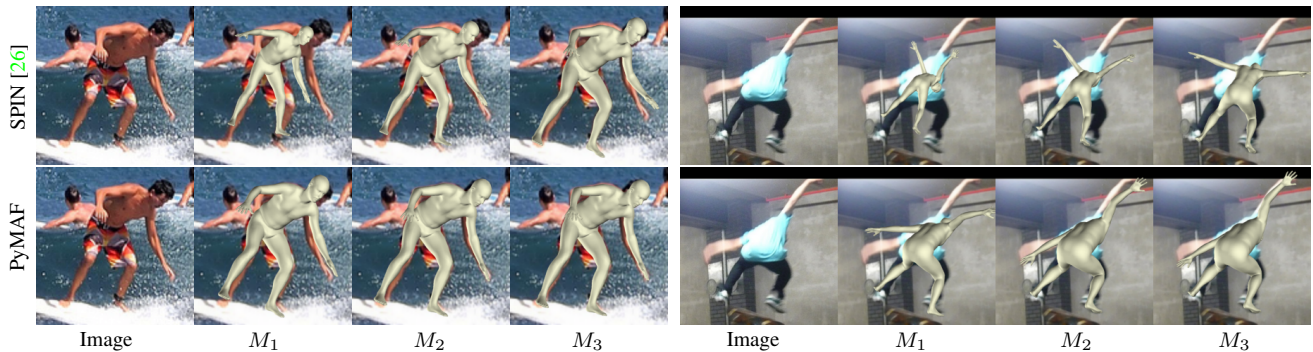}
	}\\
	\begin{tikzpicture}[remember picture,overlay]
	\node[font=\fontsize{8pt}{8pt}\selectfont] at (-7.5,0.2) {Image};
	\node[font=\fontsize{8pt}{8pt}\selectfont] at (-5.4,0.2) {$M_1$};
 	\node[font=\fontsize{8pt}{8pt}\selectfont] at (-3.2,0.2) {$M_2$};
	\node[font=\fontsize{8pt}{8pt}\selectfont] at (-1.1,0.2) {$M_3$};
	\node[font=\fontsize{8pt}{8pt}\selectfont] at (1.1,0.2) {Image};
	\node[font=\fontsize{8pt}{8pt}\selectfont] at (3.2,0.2) {$M_1$};
 	\node[font=\fontsize{8pt}{8pt}\selectfont] at (5.4,0.2) {$M_2$};
	\node[font=\fontsize{8pt}{8pt}\selectfont] at (7.5,0.2) {$M_3$};
	\end{tikzpicture}
	\vspace{-2mm}
	\caption{Qualitative differences between each iterative loop of the SPIN~\cite{kolotouros2019learning} using global features \vs the PyMAF using spatial features.}
	\label{fig:ief_maf_loop}
\end{figure*}

\begin{figure*}[t]
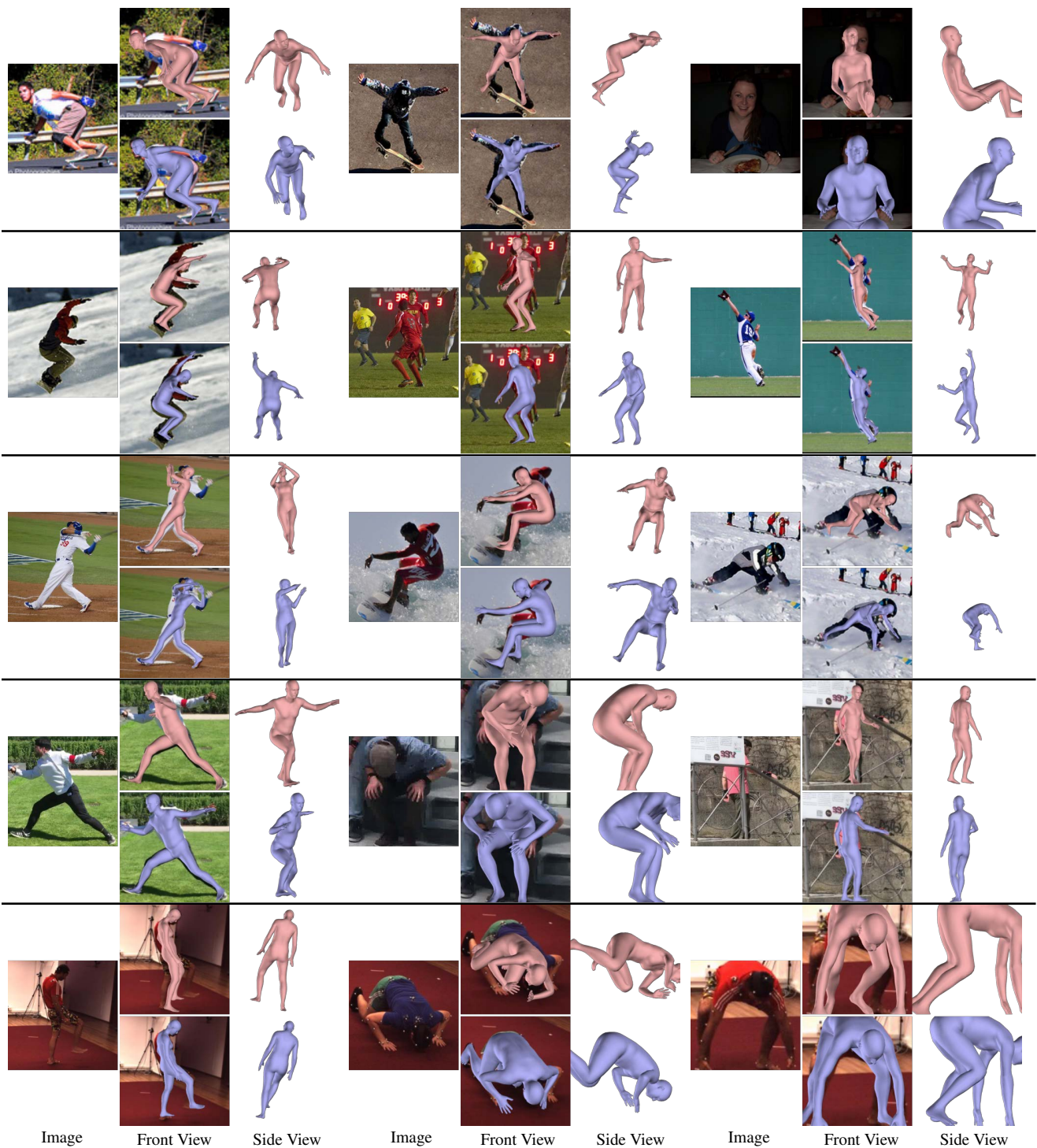

	\centering
% 	\hspace{1mm}
% 	\hspace{0.5mm}
	\foreach \idx in {1, 2, 3} {
	    \includegraphics[width=0.32\textwidth]{fig/vis/view/cpr_2view_\idx.pdf}
	    \vspace{-1.85mm}
	}
	\vspace{1.5mm}
	\noindent\rule{\textwidth}{1pt}
	\foreach \idx in {4, 5, 6} {
	    \includegraphics[width=0.32\textwidth]{fig/vis/view/cpr_2view_\idx.pdf}
	    \vspace{-1.85mm}
	}
	\vspace{1.5mm}
	\noindent\rule{\textwidth}{1pt}
	\foreach \idx in {7, 8, 9} {
	    \includegraphics[width=0.32\textwidth]{fig/vis/view/cpr_2view_\idx.pdf}
	    \vspace{-1.85mm}
	}
	\vspace{1.5mm}
	\noindent\rule{\textwidth}{1pt}
	\foreach \idx in {11, 12, 13} {
	    \includegraphics[width=0.32\textwidth]{fig/vis/view/cpr_2view_\idx.pdf}
	    \vspace{-1.85mm}
	}
	\vspace{1.5mm}
	\noindent\rule{\textwidth}{1pt}
	\foreach \idx in {21, 22, 23} {
	    \includegraphics[width=0.32\textwidth]{fig/vis/view/cpr_2view_\idx.pdf}
	}
	\\
	\begin{tikzpicture}[remember picture,overlay]
	\node[font=\fontsize{8pt}{8pt}\selectfont] at (-7.6,0.2) {Image};
	\node[font=\fontsize{8pt}{8pt}\selectfont] at (-5.7,0.2) {Front View};
 	\node[font=\fontsize{8pt}{8pt}\selectfont] at (-3.8,0.2) {Side View};
	\node[font=\fontsize{8pt}{8pt}\selectfont] at (-1.7,0.2) {Image};
	\node[font=\fontsize{8pt}{8pt}\selectfont] at (0.1,0.2) {Front View};
 	\node[font=\fontsize{8pt}{8pt}\selectfont] at (2.0,0.2) {Side View};
	\node[font=\fontsize{8pt}{8pt}\selectfont] at (4.0,0.2) {Image};
	\node[font=\fontsize{8pt}{8pt}\selectfont] at (5.9,0.2) {Front View};
 	\node[font=\fontsize{8pt}{8pt}\selectfont] at (7.8,0.2) {Side View};
	\end{tikzpicture}
	\vspace{-2mm}
	\caption{Qualitative comparison of the reconstruction results between SPIN~\cite{kolotouros2019learning} and our PyMAF approach. For each example, the upper / lower results correspond to the reconstructed meshes of SPIN (pink) / PyMAF (purple). Examples come from various datasets, including COCO (Rows 1-3), 3DPW (Row 4), and Human3.6M (Row 5).}
% 	\vspace{-3mm}
	\label{fig:view}
\end{figure*}

\textbf{More Qualitative Results.} We provide more qualitative results and compare our PyMAF with the state-of-the-art approach SPIN~\cite{kolotouros2019learning}.
Figure~\ref{fig:ief_maf_loop} shows the qualitative differences between each iterative loop in SPIN~\cite{kolotouros2019learning} and PyMAF, which uses the global features and spatial features for the parameter update respectively.
We can see that PyMAF convergences much faster and corrects the mesh parameters more effectively.
In Figure~\ref{fig:view}, we show more reconstruction results for qualitative comparisons with SPIN on both indoor and in-the-wild datasets, where PyMAF can produce natural results which are better-aligned with the images under challenging cases.
Note that our approach is complementary to SPIN, since SPIN aims at providing better supervision for the regression network, while our work focuses on the architecture design of the regression network.

\begin{figure*}[t]
	\centering
% 	\hspace{1mm}
% 	\hspace{0.5mm}
	\foreach \idx in {1, 2, 3, 4, 5, 6, 7, 8, 9, 10, 11, 12, 13, 14, 15, 16} {
	    \includegraphics[width=0.48\textwidth]{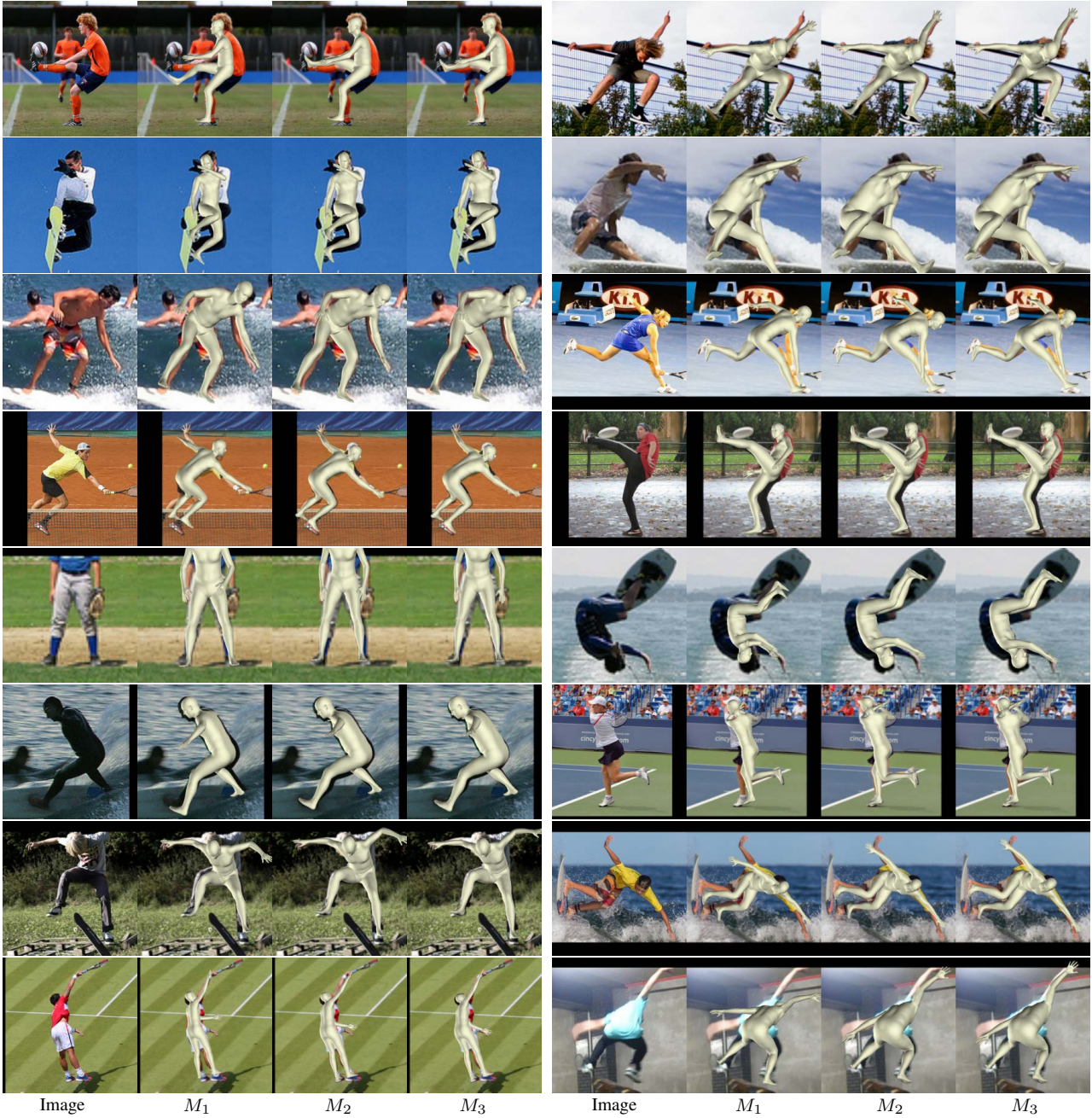}
	}\\
	\begin{tikzpicture}[remember picture,overlay]
	\node[font=\fontsize{8pt}{8pt}\selectfont] at (-7.5,0.2) {Image};
	\node[font=\fontsize{8pt}{8pt}\selectfont] at (-5.4,0.2) {$M_1$};
 	\node[font=\fontsize{8pt}{8pt}\selectfont] at (-3.2,0.2) {$M_2$};
	\node[font=\fontsize{8pt}{8pt}\selectfont] at (-1.1,0.2) {$M_3$};
	\node[font=\fontsize{8pt}{8pt}\selectfont] at (1.1,0.2) {Image};
	\node[font=\fontsize{8pt}{8pt}\selectfont] at (3.2,0.2) {$M_1$};
 	\node[font=\fontsize{8pt}{8pt}\selectfont] at (5.4,0.2) {$M_2$};
	\node[font=\fontsize{8pt}{8pt}\selectfont] at (7.5,0.2) {$M_3$};
	\end{tikzpicture}
	\vspace{-2mm}
	\caption{Successful results of PyMAF on the COCO dataset. For each example from left to right: image, the results after each iteration.}
% 	\vspace{-3mm}
	\label{fig:loop_coco}
\end{figure*}

\begin{figure*}[t]
	\centering
% 	\hspace{1mm}
% 	\hspace{0.5mm}
	\foreach \idx in {1, 2, 3, 4, 5, 6, 7, 8} {
	    \includegraphics[width=0.48\textwidth]{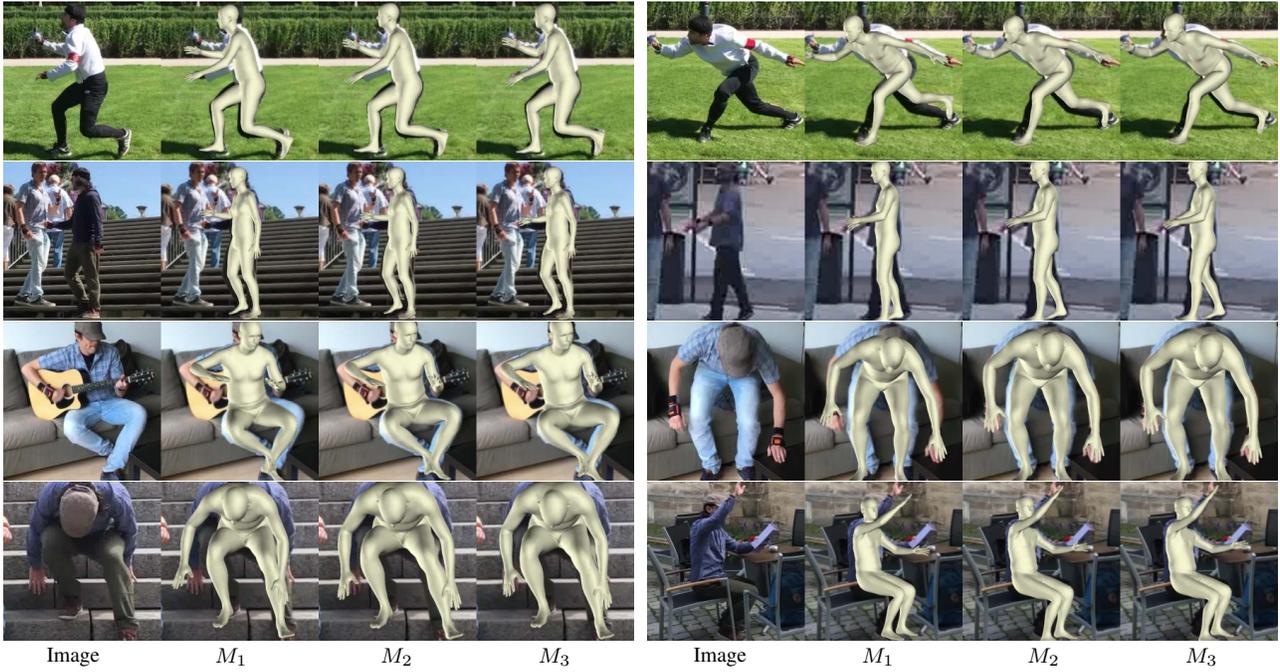}
	}
	\\
	\begin{tikzpicture}[remember picture,overlay]
	\node[font=\fontsize{8pt}{8pt}\selectfont] at (-7.5,0.2) {Image};
	\node[font=\fontsize{8pt}{8pt}\selectfont] at (-5.4,0.2) {$M_1$};
 	\node[font=\fontsize{8pt}{8pt}\selectfont] at (-3.2,0.2) {$M_2$};
	\node[font=\fontsize{8pt}{8pt}\selectfont] at (-1.1,0.2) {$M_3$};
	\node[font=\fontsize{8pt}{8pt}\selectfont] at (1.1,0.2) {Image};
	\node[font=\fontsize{8pt}{8pt}\selectfont] at (3.2,0.2) {$M_1$};
 	\node[font=\fontsize{8pt}{8pt}\selectfont] at (5.4,0.2) {$M_2$};
	\node[font=\fontsize{8pt}{8pt}\selectfont] at (7.5,0.2) {$M_3$};
	\end{tikzpicture}
	\vspace{-2mm}
	\caption{Successful results of PyMAF on the 3DPW dataset. Examples have the same layout with Figure~\ref{fig:loop_coco}.}
% 	\vspace{-3mm}
	\label{fig:loop_3dpw}
\end{figure*}

\begin{figure*}[t]
	\centering
% 	\hspace{1mm}
% 	\hspace{0.5mm}
	\foreach \idx in {1, 2, 3, 4, 5, 6} {
	    \includegraphics[width=0.48\textwidth]{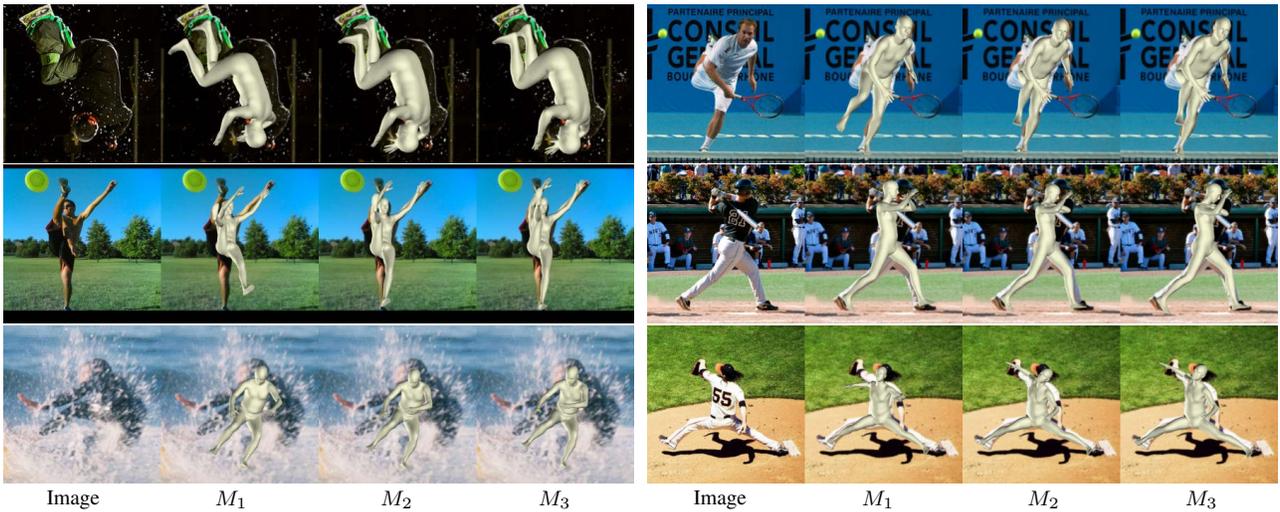}
	}
	\\
	\begin{tikzpicture}[remember picture,overlay]
	\node[font=\fontsize{8pt}{8pt}\selectfont] at (-7.5,0.2) {Image};
	\node[font=\fontsize{8pt}{8pt}\selectfont] at (-5.4,0.2) {$M_1$};
 	\node[font=\fontsize{8pt}{8pt}\selectfont] at (-3.2,0.2) {$M_2$};
	\node[font=\fontsize{8pt}{8pt}\selectfont] at (-1.1,0.2) {$M_3$};
	\node[font=\fontsize{8pt}{8pt}\selectfont] at (1.1,0.2) {Image};
	\node[font=\fontsize{8pt}{8pt}\selectfont] at (3.2,0.2) {$M_1$};
 	\node[font=\fontsize{8pt}{8pt}\selectfont] at (5.4,0.2) {$M_2$};
	\node[font=\fontsize{8pt}{8pt}\selectfont] at (7.5,0.2) {$M_3$};
	\end{tikzpicture}
	\vspace{-2mm}
	\caption{Erroneous reconstructions of our network. Though PyMAF can improve the alignment of some body parts, it remains challenging for PyMAF to correct those body parts with severe deviations, heavy occlusions, or ambiguous limb connections. Examples have the same layout with Figure~\ref{fig:loop_coco}.}
	\label{fig:failed_cases}
\end{figure*}

To demonstrate the efficacy of the parameter rectification in PyMAF, we further provide more examples on COCO and 3DPW in Figures~\ref{fig:loop_coco} and \ref{fig:loop_3dpw}, respectively.
We can observe that PyMAF improves the mesh-image alignment progressively by correcting the predictions based on the current observations.
We also visualize some erroneous results of our approach in Figure~\ref{fig:failed_cases}, where PyMAF may fail when the initial reconstructed results have severe deviations due to the heavy occlusions or ambiguous limb connections in complex scenes.

% \subsection{Further Discussions}
% \label{sec:discuss}

% The proposed PyMAF is primarily motivated by the observation of the reprojection misalignment of parametric mesh results.
% Though PyMAF improves the mesh-image alignment on 2D image planes, it can still hardly address the depth ambiguity problem in 3D space.
% Recently, there have also been approaches to improving the alignment in 3D space, which achieve impressive performances on 3D pose estimation benchmarks.
% These approaches involve 3D skeleton estimation in their pipelines and obtain the final meshes by calibration~\cite{luan2021pc} or inverse kinematics~\cite{li2020hybrik}.
% In the future, we would like to combine the merits of the approaches in these two directions.

\end{document}